\documentclass[11pt]{article}

\usepackage[preprint]{acl}

\usepackage{times}
\usepackage{latexsym}

\usepackage[T1]{fontenc}

\usepackage[utf8]{inputenc}

\usepackage{microtype}
\usepackage{booktabs}

\usepackage{inconsolata}

\usepackage{graphicx}

  \usepackage[table]{xcolor}
  \definecolor{agreeLow}{RGB}{229,245,224}
  \definecolor{agreeMid}{RGB}{116,196,118}
   \definecolor{agreeHigh}{RGB}{35,139,69}

  \definecolor{werLow}{RGB}{222,235,247}
  \definecolor{werMid}{RGB}{107,174,214}
  \definecolor{werHigh}{RGB}{33,113,181}

\usepackage{pifont}
\usepackage{xcolor}
\usepackage{xspace}
\usepackage{enumitem}
\usepackage{bm}
\usepackage{todonotes}
\usepackage{xcolor}
\usepackage{multirow}
\usepackage{float}
\usepackage{comment}

\newcommand*{\corpus}{\texttt{OpenBibleTTS}\xspace}

\newcommand{\cmark}{\textcolor{green}{\ding{51}}}
\newcommand{\xmark}{\textcolor{red}{\ding{55}}}

\title{OpenBibleTTS: Large-Scale Speech Resources and TTS Models for Low-Resource Languages}

\author{
 \textbf{David Guzmán\textsuperscript{1,2}}\quad
 \textbf{Luel Hagos Beyene\textsuperscript{3,4}\thanks{Work done during internship at McGill and Mila}}\quad
 \textbf{Jesujoba Oluwadara Alabi\textsuperscript{5}}
\\
 \textbf{Yejin Jeon\textsuperscript{1,2}}\quad
 \textbf{Dietrich Klakow\textsuperscript{5}}\quad
 \textbf{David Ifeoluwa Adelani\textsuperscript{1,2,6}}
\\
\\
 \textsuperscript{1}McGill University\quad
 \textsuperscript{2}Mila - Quebec AI Institute\quad
 \textsuperscript{3}AIMS Research and Innovation Centre\\
 \textsuperscript{4}NM-AIST\quad
 \textsuperscript{5}Saarland University\quad
 \textsuperscript{6}Canada CIFAR AI Chair
\\
 \small{
   \textbf{Correspondence:} 
   \texttt{\{david.guzman, david.adelani\}@mila.quebec}
 }
}

\begin{document}
\maketitle

\begin{abstract}
Recent advances in neural text-to-speech (TTS) and multilingual speech generation have substantially improved synthetic speech quality, yet these gains remain unevenly distributed across the world's languages. Existing models are still dominated by a small set of high-resource languages, while many studies of low-resource TTS are simulated on artificially downsampled high-resource corpora that do not reflect the orthographic variation and limited phonetic coverage encountered in genuinely underrepresented settings. As such, we introduce OpenBibleTTS, which is a large-scale benchmark for low-resource speech synthesis spanning 37 underrepresented languages. Moreover, a systematic comparison of various TTS architectures and large-scale speech generation models is conducted across in-domain Biblical text and out-of-domain material. Results show that no single system dominates across languages and metrics: Gemini-TTS achieves the highest listener ratings on most evaluated languages, but monolingual EveryVoice models trained on OpenBibleTTS remain strongest for intelligibility and are preferred in several African languages, while open from-scratch systems degrade sharply on out-of-domain text, revealing a persistent gap between broad multilingual coverage and reliable synthesis quality in underserved linguistic communities. We complement automatic evaluation with subjective human judgments, and open-source all processed datasets, alignments, and trained models to support future low-resource TTS research. 
\end{abstract}

\begin{center}
\small
\textbf{Code \& data:}
\href{https://huggingface.co/multilingual-tts}{Datasets \& models} $\cdot$
\href{https://github.com/McGill-NLP/open-bible-resources}{Alignment pipeline} $\cdot$
\href{https://github.com/McGill-NLP/open-bible-tts}{Training code}
\end{center}

\section{Introduction}

Advances in neural text-to-speech (TTS) systems \cite{ren2021fastspeech2, kim2021vits} have led to substantial improvements in synthetic speech naturalness and intelligibility. Building upon these developments, recent research has increasingly shifted focus toward multilingual speech generation, with the aim of extending high-quality TTS capabilities beyond a single language \cite{pmlr-v162-casanova22a, zhang2023speakforeignlanguagesvoice, jiang2024megatts}. Nevertheless, even though more than 7,000 languages are spoken worldwide, research remain overwhelmingly centered on a small set of high-resource languages (HRLs); YourTTS \citep{pmlr-v162-casanova22a} covers three languages, whereas Valle-X \citep{zhang2023speakforeignlanguagesvoice} and MegaTTS2 \citep{jiang2024megatts} cover the two languages of English and Chinese. Consequently, the robustness and generalization ability of contemporary TTS systems for genuinely underrepresented languages remain insufficient.

A major barrier in progress for such low-resource settings is the lack of publicly available resources. Unlike HRLs, where large curated datasets \citep{park2019css10collectionsinglespeaker, ardila-etal-2020-common, koizumi2023librittsrrestoredmultispeakertexttospeech, Cmltts2023} have enabled rapid model development \citep{casanova24_interspeech}, many underrepresented languages still lack sufficiently paired speech-text data, preprocessing tools, and open-source pipelines. As a result, existing studies rely on simulated low-resource settings, where HRLs are artificially downsampled to smaller training subsets \citep{lux-etal-2022-low, Simulated10444075, kim-etal-2024-audio}. Consequently, they fail to capture the practical challenges inherent to genuinely low-resource languages (LRLs), including orthographic variation, limited phonetic coverage, and noisy alignments. 

For many LRLs, the most available speech data often come from the religious domain, such as the Bible, which has led to a growing number of works leveraging these corpora to build single-speaker TTS models, including CMU Wilderness~\citep{Black2019CMUWM} and BibleTTS~\citep{meyer22c_interspeech}. The main difficulty in further scaling to more languages now lies in the availability of permissive licenses for releasing TTS data and/or the challenge of correctly aligning speech data with text pairs.

In this paper, we introduce \corpus, a new training and evaluation dataset for 37 LRLs curated from public available Bible Speech corpus with open license.\footnote{\url{https://www.open.bible/}} 
Our dataset is first constructed by processing and aligning audio recordings with corresponding textual transcripts using timing metadata and forced alignment. Leveraging this dataset, we  evaluate both conventional neural TTS architectures and recent large-scale speech generation systems (e.g., EveryVoice, VITS, F5 \citep{chen-etal-2025-f5}, OmniVoice \citep{zhu2026omnivoice}, and Gemini-TTS 2.5 Pro. To systematically evaluate the various TTS paradigms, we ask the following 
research questions: (1) how different TTS architectures behave across languages and the effect of data size, (2) whether large-scale speech foundation models substantially outperform or potentially underperform models that are trained from scratch, and (3) how synthesis quality and intelligibility change when moving from in-domain to out-of-domain settings. 

Beyond in-domain evaluation on Biblical text \cite{openbible}, we further investigate out-of-domain generalization using sentences from Fleurs-Wikipedia~\citep{goyal-etal-2022-flores} and Bouquet-conversational sentences~\citep{andrews-etal-2025-bouquet}.
Overall, our work provides a clearer understanding of the current capabilities and limitations of modern TTS systems for genuinely low-resource languages, and the remaining challenges toward more inclusive and globally accessible speech technologies. To support future low-resource research, we publicly release the processed audio datasets, alignments, and trained models for all 37 languages.


\begin{figure*}[ht!]
    \centering
    \includegraphics[width=0.8\linewidth]{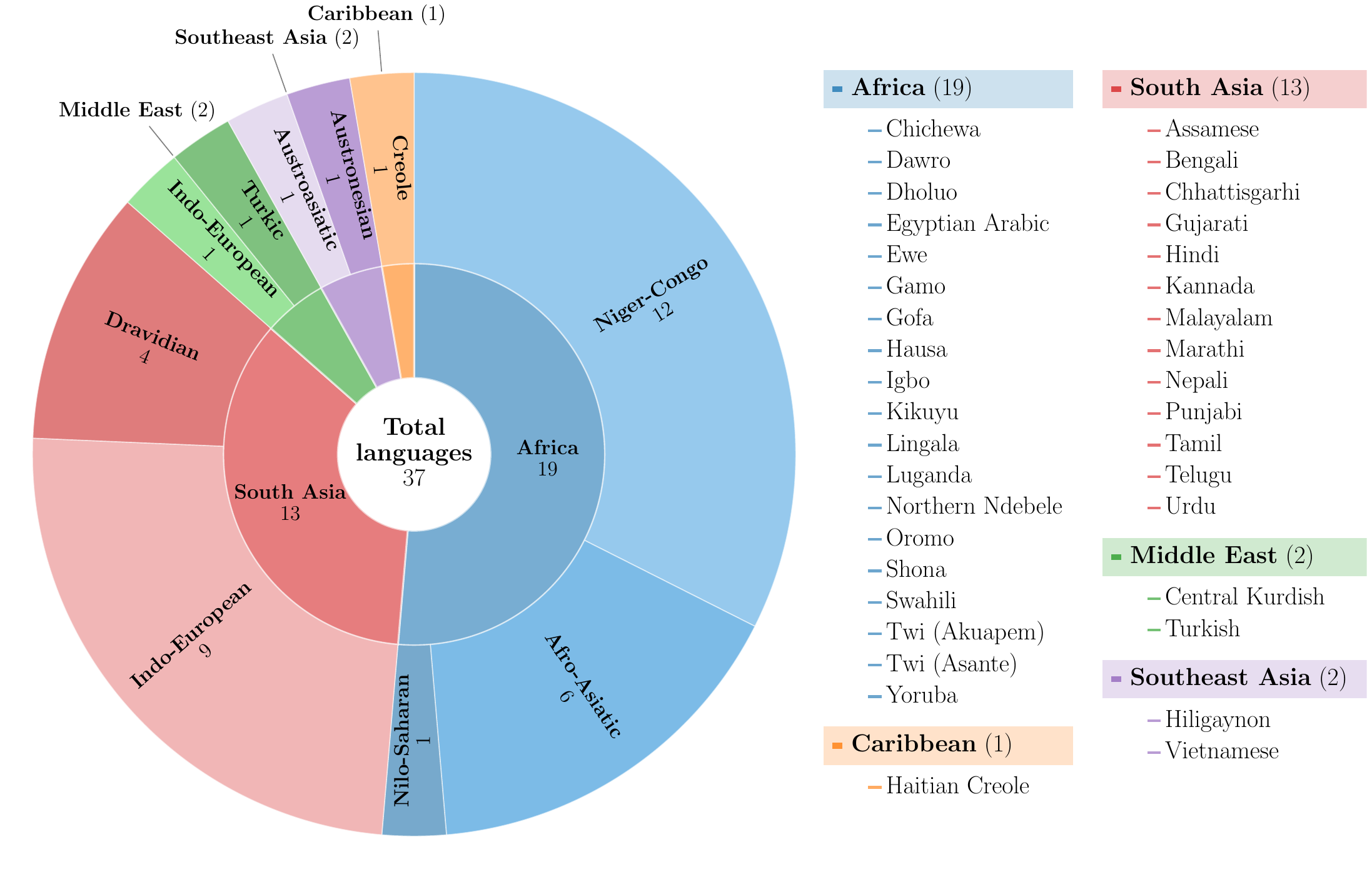}
    \caption{Geographic and genealogical coverage of the 37 languages in \corpus. The inner ring partitions languages by region (Africa, South Asia, Middle East, Southeast Asia, Caribbean) and the outer ring further breaks each region down by language family.}
    \label{fig:sunburst_languages}
\end{figure*}

\section{Related Work}
\paragraph{Multilingual Speech Datasets} The development of high-quality speech datasets has been critical for advances in speech technologies, particularly TTS~\citep{Tan2021ASO}. However, many existing corpora, including multilingual TTS datasets such as Multilingual MLS~\citep{Pratap2020MLSAL} and CML-TTS~\citep{Cmltts2023}, remain dominated by English and other high-resource languages. Broader multilingual speech datasets such as Mozilla Common Voice~\citep{ardila-etal-2020-common},  FLEURS~\citep{Conneau2022FLEURSFL}, and \cite{pratap2023scalingspeechtechnology1000} have substantially expanded language coverage, but were primarily designed for ASR and do not release trained artefacts. Although these datasets have been repurposed for TTS, such as in FLEURS-R~\citep{ma24c_interspeech}, prior work has reported data quality issues in the original corpora~\citep{lau-etal-2025-data,alabi25_interspeech}.

Prior work has leveraged aligned multilingual text sources such as the Bible for constructing TTS corpora, including CMU Wilderness~\citep{Black2019CMUWM}, which provides New Testament speech data for approximately 700 languages, as well as AfricanVoices~\citep{ogayo22_interspeech} and BibleTTS~\citep{meyer22c_interspeech}, which focus on African languages. Our work builds on this line of research by adopting the same aligned biblical text paradigm to construct a larger multilingual speech dataset covering languages across diverse regions.


\vspace{-8pt}
\paragraph{Neural TTS Architectures}
Early TTS systems were based on statistical parametric approaches, which relied on handcrafted pipelines for speech synthesis~\citep{Zen2007StatisticalPS,Tan2021ASO}. While these systems produced intelligible speech, they lacked natural prosody and required extensive feature engineering. The introduction of neural architectures such as Tacotron~\citep{wang17n_interspeech} significantly advanced TTS quality ~\citep{Tan2021ASO} by directly mapping text to mel-spectrograms, though they often suffered from slow inference and robustness issues. In parallel, neural vocoders such as WaveNet~\citep{vandenoord16_ssw}, WaveGlow~\citep{Prenger2018WaveglowAF}, and HiFi-GAN~\citep{Kong2020HiFiGANGA} improved waveform reconstruction.

Subsequent architectures such as FastSpeech~\citep{ren2021fastspeech2} and VITS~\citep{kim2021vits} improved synthesis speed, stability, and speech naturalness through non-autoregressive and variational modeling approaches. More recent diffusion- and flow-matching-based systems such as E2 TTS~\citep{Eskimez2024E2TE} and F5-TTS~\citep{chen-etal-2025-f5}, have further improved speech quality and generation robustness while simplifying the synthesis pipeline. In parallel, multilingual~\citep{lux-etal-2022-low,lux24_interspeech,casanova24_interspeech,Gong2023ZMMTTSZM} and pretrained large-scale speech models, particularly LLM-based and multimodal systems~\citep{Hu2026Qwen3TTSTR}, have enabled cross-lingual and zero-shot TTS synthesis via shared speaker and language representations. However, their performance across languages remains underexplored, particularly for low-resource languages. We address this by evaluating various TTS architectures across diverse languages. 


\begin{table*}[t]
\centering
\resizebox{\textwidth}{!}{%
\begin{tabular}{@{}cccccccccc@{}}
\toprule
\textbf{Language} & \multicolumn{1}{c}{\textbf{ISO 639-3}} & \textbf{Region} & \textbf{Country} & \textbf{Language Family} & \textbf{Script} & \multicolumn{1}{c}{\textbf{Joshi Class}} & \textbf{Timing Metadata} & \textbf{Total Hours} & \textbf{Total utterances} \\ \midrule
Egyptian Arabic   & arb                                    & Africa          & Egypt            & Afro-Asiatic             & Arabic          & 3                                        & \xmark \ No                          & 85.23                & 30,510                     \\
Assamese          & asm                                    & South Asia      & India            & Indo-European            & Bengali         & 1                                        & \cmark \ Yes                         & 104.49               & 30,530                     \\
Bengali           & ben                                    & South Asia      & India            & Indo-European            & Bengali         & 3                                        & \cmark \ Yes                         & 98.02                & 30,535                     \\
Central Kurdish   & ckb                                    & Middle East     & Kurdistan        & Indo-European            & Arabic          & 1                                        & \cmark \ Yes                         & 85.40                & 30,565                     \\
Chhattisgarhi     & hne                                    & South Asia      & India            & Indo-European            & Devanagari      & -                                        & \cmark \ Yes                         & 101.61               & 30,451                     \\
Chichewa          & nya                                    & Africa          & Malawi           & Niger-Congo              & Latin           & 1                                        & \xmark \ No                          & 110.98               & 30,322                     \\
Dawro             & dwr                                    & Africa          & Ethiopia         & Afro-Asiatic             & Latin           & -                                        & \xmark \ No                          & 116.98               & 29,579                     \\
Dholuo            & luo                                    & Africa          & Kenya            & Nilo-Saharan             & Latin           & -                                        & \cmark \ Yes                         & 76.02                & 30,429                     \\
Ewe               & ewe                                    & Africa          & Ghana            & Niger-Congo              & Latin           & 1                                        & \cmark \ Yes                         & 95.86                & 30,164                     \\
Gamo              & gmv                                    & Africa          & Ethiopia         & Afro-Asiatic             & Latin           & 0                                        & \cmark \ Yes                         & 107.07               & 30,200                     \\
Gofa              & gof                                    & Africa          & Ethiopia         & Afro-Asiatic             & Latin           & -                                        & \xmark \ No                          & 86.78                & 30,388                     \\
Gujarati          & guj                                    & South Asia      & India            & Indo-European            & Gujarati        & 1                                        & \cmark \ Yes                         & 83.09                & 30,447                     \\
Haitian Creole    & hat                                    & Caribbean       & Haiti            & Creole                   & Latin           & 0                                        & \xmark \ No                          & 104.03               & 30,597                     \\
Hausa             & hau                                    & Africa          & Nigeria          & Afro-Asiatic             & Latin           & 2                                        & \cmark \ Yes                         & 99.60                & 30,717                     \\
Hiligaynon        & hil                                    & Southeast Asia  & Philippines      & Austronesian             & Latin           & 0                                        & \cmark \ Yes                         & 106.43               & 29,260                     \\
Hindi             & hin                                    & South Asia      & India            & Indo-European            & Devanagari      & 4                                        & \cmark \ Yes                         & 99.09                & 30,438                     \\
Igbo              & ibo                                    & Africa          & Nigeria          & Niger-Congo              & Latin           & 1                                        & \cmark \ Yes                         & 94.46                & 30,011                     \\
Kannada           & kan                                    & South Asia      & India            & Dravidian                & Kannada         & 1                                        & \cmark \ Yes                         & 104.86               & 30,495                     \\
Kikuyu            & kik                                    & Africa          & Kenya            & Niger-Congo              & Latin           & 1                                        & \xmark \ No                          & 87.08                & 30,722                     \\
Lingala           & lin                                    & Africa          & Congo            & Niger-Congo              & Latin           & 1                                        & \cmark \ Yes                         & 128.49               & 28,790                     \\
Luganda           & lug                                    & Africa          & Uganda           & Niger-Congo              & Latin           & 1                                        & \cmark \ Yes                         & 101.75               & 30,440                     \\
Malayalam         & mal                                    & South Asia      & India            & Dravidian                & Malayalam       & 1                                        & \cmark \ Yes                         & 86.12                & 30,190                     \\
Marathi           & mar                                    & South Asia      & India            & Indo-European            & Devanagari      & 2                                        & \cmark \ Yes                         & 92.35                & 30,573                     \\
Northern Ndebele  & nde                                    & Africa          & Zimbabwe         & Niger-Congo              & Latin           & 0                                        & \cmark \ Yes                         & 100.76               & 30,161                     \\
Nepali            & nep                                    & South Asia      & Nepal            & Indo-European            & Devanagari      & 1                                        & \cmark \ Yes                         & 106.63               & 30,347                     \\
Oromo             & orm                                    & Africa          & Ethiopia         & Afro-Asiatic             & Latin           & 1                                        & \cmark \ Yes                         & 99.16                & 30,413                     \\
Punjabi           & pan                                    & South Asia      & India            & Indo-European            & Gurmukhi        & 2                                        & \cmark \ Yes                         & 86.16                & 30,496                     \\
Shona             & sna                                    & Africa          & Zimbabwe         & Niger-Congo              & Latin           & 1                                        & \xmark \ No                          & 74.66                & 30,685                     \\
Swahili           & swh                                    & Africa          & Kenya            & Niger-Congo              & Latin           & 2                                        & \xmark \ No                          & 96.38                & 30,634                     \\
Tamil             & tam                                    & South Asia      & India            & Dravidian                & Tamil           & 3                                        & \cmark \ Yes                         & 92.93                & 30,516                     \\
Telugu            & tel                                    & South Asia      & India            & Dravidian                & Telugu          & 1                                        & \cmark \ Yes                         & 93.18                & 30,059                     \\
Turkish           & tur                                    & Middle East     & Turkey           & Turkic                   & Latin           & 4                                        & \xmark \ No                          & 63.50                & 29,747                     \\
Twi (Akuapem)     & twi                                    & Africa          & Ghana            & Niger-Congo              & Latin           & 1                                        & \cmark \ Yes                         & 71.20                & 30,270                     \\
Twi (Asante)      & twi                                    & Africa          & Ghana            & Niger-Congo              & Latin           & 1                                        & \cmark \ Yes                         & 78.87                & 30,565                     \\
Urdu              & urd                                    & South Asia      & India            & Indo-European            & Arabic          & 3                                        & \cmark \ Yes                         & 88.26                & 30,634                     \\
Vietnamese        & vie                                    & Southeast Asia  & Vietnam          & Austroasiatic            & Latin           & 4                                        & \cmark \ Yes                         & 71.62                & 30,451                     \\
Yoruba            & yor                                    & Africa          & Nigeria          & Niger-Congo              & Latin           & 2                                        & \cmark \ Yes                         & 89.91                & 30,625                     \\ \midrule
\textbf{Total}    &                                        &                 &                  &                          &                 &                                          &                                      & \textbf{3,469.01}             & \textbf{1,121,956}                  \\ \bottomrule
\end{tabular}%
}
\caption{Overview of the 37 \corpus languages. ISO~639-3 code, geographic region, primary country of use, language family, writing system, Joshi resource class~\citep{joshi-etal-2020-state} (``--'' indicates languages not covered in the original taxonomy), whether timing metadata was available (otherwise force-alignment was conducted), the total aligned speech duration in hours, and the total number of utterances retained after quality filtering, is reported.}
\label{tab:language-stats}
\end{table*}

\section{OpenBibleTTS Corpus}

\subsection{Data source}
\corpus is derived from the CC BY-SA licensed Open Bible platform \citep{openbible}, a large multilingual repository of scripture produced by professional Bible-translation organizations. We use this source because it offers a rare combination of properties that are especially valuable for low-resource TTS benchmarking: long-form read speech, carefully edited transcripts, broad language coverage, consistent narrative style, and near-parallel textual content across languages.

For each language, Open Bible exposes two complementary artefact types. \textbf{Audio} is distributed as full-Bible recordings, typically organized as chapter-level MP3 files under Old-Testament and New-Testament directory structures. \textbf{Text} is distributed as scripture transcripts in Unified Standard Format Markers (USFM) and its XML counterpart (USX), both of which are widely used in professional Bible translation workflows. While this pairing gives each language a large collection of chapter-level speech files and structured verse-level text, it does not directly provide the utterance-level segmentation required for training and evaluating modern TTS systems.

Our corpus construction therefore treats Open Bible as a multilingual raw-resource layer and converts it into a TTS-ready benchmark through systematic text parsing, audio segmentation, and verse-level alignment. In doing so, OpenBibleTTS is inspired by the work of \citet{meyer22c_interspeech}, that showed that high-fidelity scripture audio can serve as a reproducible testbed for speech synthesis, while extending that paradigm in two directions central to this paper: from a primarily African-language setting to 37 languages across five geographic regions, and from corpus creation alone to a controlled comparison of from-scratch neural TTS architectures and proprietary large-scale TTS models under the same evaluation conditions.

\subsection{Languages}
Figure \ref{fig:sunburst_languages} and Table \ref{tab:language-stats} provide an overview of the 37 languages in OpenBibleTTS, selected to capture substantial geographic, genealogical, and orthographic diversity among underrepresented speech communities. The corpus spans five regions: \textbf{Africa} (19 languages), \textbf{South Asia} (13), \textbf{Southeast Asia} (2), the \textbf{Middle East} (2), and the \textbf{Caribbean} (1). It further covers nine language families of Niger-Congo, Indo-European, Afro-Asiatic, Dravidian, Turkic, Austronesian, Austroasiatic, Nilo-Saharan, and Creole, as well as seven writing systems that include Latin (22 languages), Devanagari (4), Arabic (3), Dravidian scripts (Kannada, Malayalam, Tamil, and Telugu; 4 in total), Bengali (2), Gujarati (1), and Gurmukhi (1).
To contextualise these languages within the broader resource landscape, we additionally assign each language a Joshi resource class following \citet{joshi-etal-2020-state}. 

\subsection{Text-Audio Alignment}
The two alignment pipelines described below produce standardized verse-level speech--text pairs, which consist of a segmented waveform that is saved as \texttt{\small\{BOOK\}\_\{CHAPTER\}\_\{VERSE\}.wav} and a paired \texttt{.txt} file containing the original transcript.

\paragraph{Timing-file Alignment (28 languages).}
For languages with timing metadata, we first parse the USFM/USX scripture files to extract canonical verse text, then cross-reference each verse boundary against the word-level timestamps provided in the corresponding timing JSON.
The chapter-level MP3 recordings are subsequently cut at these verse boundaries using \texttt{ffmpeg}, producing mono WAV clips. To remove non-scriptural preambles, such as narrator announcements, we discard all audio preceding the timestamp of the first scripted word.

\paragraph{Forced Alignment (9 languages).}
For languages without timing files, we use the zero-shot \textbf{ReadAlongs Studio} \citep{littell-etal-2022-readalong} forced aligner that takes a text transcript and a chapter-level audio file, and produces word-level boundaries stored in Praat TextGrid.
The aligner internally converts text into phonemes via a grapheme-to-phoneme (g2p) back-end and optimizes the alignment using CTC-based dynamic programming.
Since many low-resource languages lack a dedicated g2p model, the language code \texttt{und} (\emph{undetermined}) is used as a universal fallback, which maps unseen graphemes to their nearest phoneme equivalents.

Prior to alignment, we apply minimal text normalization to the verse transcripts, by removing numerals and punctuation marks such as colons, semicolons, and quotation marks, which the g2p back-end cannot consistently map to phoneme sequences. This normalized text is used only for alignment; the original transcript, with punctuation preserved, is retained in the paired \texttt{.txt} file released with each audio clip. To account for pre-verse narrator announcements, we prepend a dummy transcript line that absorbs the non-scriptural audio during alignment and then discard the resulting boundary. The chapter recording is finally segmented into verse-level clips, yielding the same output format as the timing-file pipeline.

\subsection{Speaker Diarization}
Several language collections contain recordings from more than one speaker. Since unlabelled speaker variation can confound voice identity and reduce speaker consistency in TTS training, we assign speaker labels before releasing the corpus.

To do this, instead of running diarization over the full corpus, we run diarization over every language by exploiting the fact that all chapters within a Bible book share the same narrator. Specifically, we concatenate the first verse of each book (up to 66 samples, separated by one second of silence) into a 10-15 minute reference file, and submit it to the \textbf{pyannote/speaker-diarization-precision-2} pipeline~\citep{bredin23_interspeech}. The speaker detected within each book's time window is recorded as that book's narrator, and the label is propagated to all verses of that book, allowing speakers to be identified and distinguished across the corpus for each language.
The resulting \texttt{speaker\_id} is included as a metadata column in the dataset.

\subsection{Data Quality Filtering and Statistics}
After alignment and diarization, we apply language-specific quality filtering following the data-quality framework of \citet{meyer22c_interspeech}. Each paired speech-text sample is evaluated against four criteria, and is assigned to the \textsc{Best} subset only if it satisfies all of them.

\begin{itemize}[noitemsep,topsep=2pt,leftmargin=1em]
  \item \textbf{Duration ceiling}: clips longer than 30 seconds are removed, since unusually long segments often indicate boundary or alignment failures.
  \item \textbf{Transcript floor}: pairs with fewer than 10 characters are discarded, as these typically correspond to section headings or metadata artefacts.
  \item \textbf{CTC feasibility}: pairs whose transcripts are disproportionately long relative to their audio duration are excluded, since such examples are ill-posed for CTC-based alignment and training.
  \item \textbf{Ratio outliers}: pairs with audio-to-text ratios more than three standard deviations from the per-language mean are filtered out.
\end{itemize}

\noindent In total, \corpus contains \textbf{3,469 hours} of aligned speech and \textbf{1,121,956 utterances} across \textbf{37 languages}.

\section{Experimental Setup}
For evaluation, we make use of the scripture-aligned recordings from \corpus{}. Each language is supplied as a multi-speaker train/test split of studio recordings at $24$~kHz, with roughly $25$--$30$~thousand training utterances per language. Per-language row counts are reported in \autoref{tab:language-stats}.

\subsection{TTS Models} 
\noindent \autoref{tab:tts-systems} summarizes the five systems in the synthesis paradigm, the provenance of the vocoder, and the components that we trained in \corpus.

\paragraph{FastSpeech~2}~\citep{ren2021fastspeech2} is a non-autoregressive acoustic model with explicit duration, pitch, and energy predictors that maps text to a mel-spectrogram which is then rendered to a waveform by a vocoder, representing the supervised cascaded paradigm. We use the \textbf{EveryVoice} toolkit \citep{pine2024everyvoice}, which is an open-source toolkit designed for low-resource TTS that pairs a modified \textbf{FastSpeech 2} acoustic model and an \textbf{iSTFTNet} vocoder \citep{kaneko2022istftnetfastlightweightmelspectrogram}. For each language, we train the FastSpeech 2 acoustic model with multispeaker embeddings from scratch, then fine-tune the pretrained vocoder.

\paragraph{VITS} ~\citep{kim2021vits} represents the fully end-to-end (E2E) paradigm: a conditional VAE with a normalizing-flow prior and adversarial training maps text directly to a waveform in a single network, with no separately-trained vocoder at inference. We train VITS end-to-end from scratch per language with learned speaker embeddings.

\paragraph{F5-TTS} ~\citep{chen-etal-2025-f5} is representative of flow-matching zero-shot voice cloning models; a Diffusion Transformer (DiT) backbone is trained to predict a mel-spectrogram from target text and a short reference audio clip. The mel-spectrogram is converted to a waveform by a off-the-shelf, pretrained Vocos vocoder \citep{siuzdak2024vocos}, which we use  without fine-tuning.

\paragraph{OmniVoice}~\citep{zhu2026omnivoice} pushes zero-shot multilingual synthesis to massive language coverage: text is directly mapped to multi-codebook acoustic tokens via a discrete non-autoregressive diffusion language model, with its bidirectional Transformer backbone initialized from the pretrained \texttt{Qwen3-0.6B-Base} LLM \citep{qwen3technicalreport}, which was trained on 581k~hours and covers more than 600 languages. A single reference clip per language is drawn from the Open Bible train split.

\paragraph{Gemini-TTS} (\texttt{gemini-2.5-pro-preview-tts}) is a closed-source commercial reference point, and is queried with the \texttt{Kore} voice per language.

\begin{table}[t]
\centering
\small
\setlength{\tabcolsep}{4pt}
\renewcommand{\arraystretch}{1.05}
\scalebox{0.88}{
\begin{tabular}{@{}cccc@{}}
\toprule
\textbf{System} & \textbf{Paradigm} & \textbf{Vocoder} & \textbf{Trained Components} \\
\midrule
EveryVoice   & Cascade       & iSTFTNet     & Acoustic + Vocoder (FT) \\
VITS         & E2E VAE       & Integrated   & Full model \\
F5-TTS       & Flow matching & Vocos (pre.) & Acoustic only \\
OmniVoice    & Diffusion LM  & Codec (pre.) & ---  \\
Gemini TTS   & Proprietary   & Undisclosed  & ---  \\
\bottomrule
\end{tabular}
}
\caption{The five TTS systems compared in this study. ``pre.''~= pretrained vocoder. ``FT''~= fine-tuned from a pretrained multilingual checkpoint.}
\vspace{-5mm}
\label{tab:tts-systems}
\end{table}


\subsection{Model Training and Evaluation }
\paragraph{Training.} 
EveryVoice, VITS, and F5-TTS models are trained monolingually from scratch for each of the 37 languages under identical settings: 500{,}000 total optimizer updates with the default architecture and learning-rate schedule from the corresponding upstream implementation. 
OmniVoice and Gemini TTS are used off-the-shelf without per-language training. Hardware configurations and full training recipes are reported in Appendix \ref{sec:appendix_training_configuration}.

\paragraph{Evaluation.}Two complementary evaluations on the same set of synthesized utterances are reported: automatic metrics across all 37 languages and a human listening study on a 10-language subset.

\paragraph{Automatic metrics.} For every (system, language) pair we score the first 500 held-out utterances from the \corpus test split. 1) \textit{Intelligibility} is reported as word error rate (WER)\footnote{\url{https://github.com/jitsi/jiwer}} between the input text and the transcription with the Omnilingual 1B ASR model \citep{omnilingualasrteam2025omnilingualasropensourcemultilingual}. When conducting ASR for WER calculation, synthesized waveforms are passed together with a Flores-style language code (e.g.\ \texttt{swh\_Latn}, \texttt{yor\_Latn}) for correct transcription. Reference and hypothesis texts are normalized with a Whisper-style pipeline before scoring; for Arabic-script languages we additionally strip combining marks after NFD decomposition. 2) \textit{Naturalness} is reported as predicted mean opinion score from UTMOSv2~\citep{baba2024utmosv2} without fine-tuning. Since UTMOSv2 was trained predominantly on English read speech, we treat its score as a relative ranking signal within each language rather than an absolute MOS estimate.

\paragraph{Human evaluation.} Ten of the 37 languages were rated by three native-speaker annotators each, chosen to include at least one language from each geographic region in \corpus{} (Africa, South Asia, Middle East, Southeast Asia, and the Caribbean). For each language, we sampled 20 utterances from the Open Bible test split, 20 from FLEURS, and 20 from BOUQuET, and synthesized every system on the same text so comparisons are content-controlled and model-blinded. Annotators judged each clip on a single 1-5 mean opinion score that jointly covers naturalness and intelligibility. This yields 340 clips per language in most cases---120 from Open~Bible (20 utterances $\times$ 6 systems), 120 from FLEURS, and 100 from BOUQuET (no ground-truth audio)---with 320 for Haitian Creole since FLEURS lacks ground-truth recordings for that language. Recruitment, compensation, and annotation-platform details are reported in Appendix~\ref{sec:appendix_human_evaluation}.

\section{Results and Discussion}

\begin{figure*}[ht!]
    \centering
    \resizebox{0.98\textwidth}{!}{%
        \includegraphics{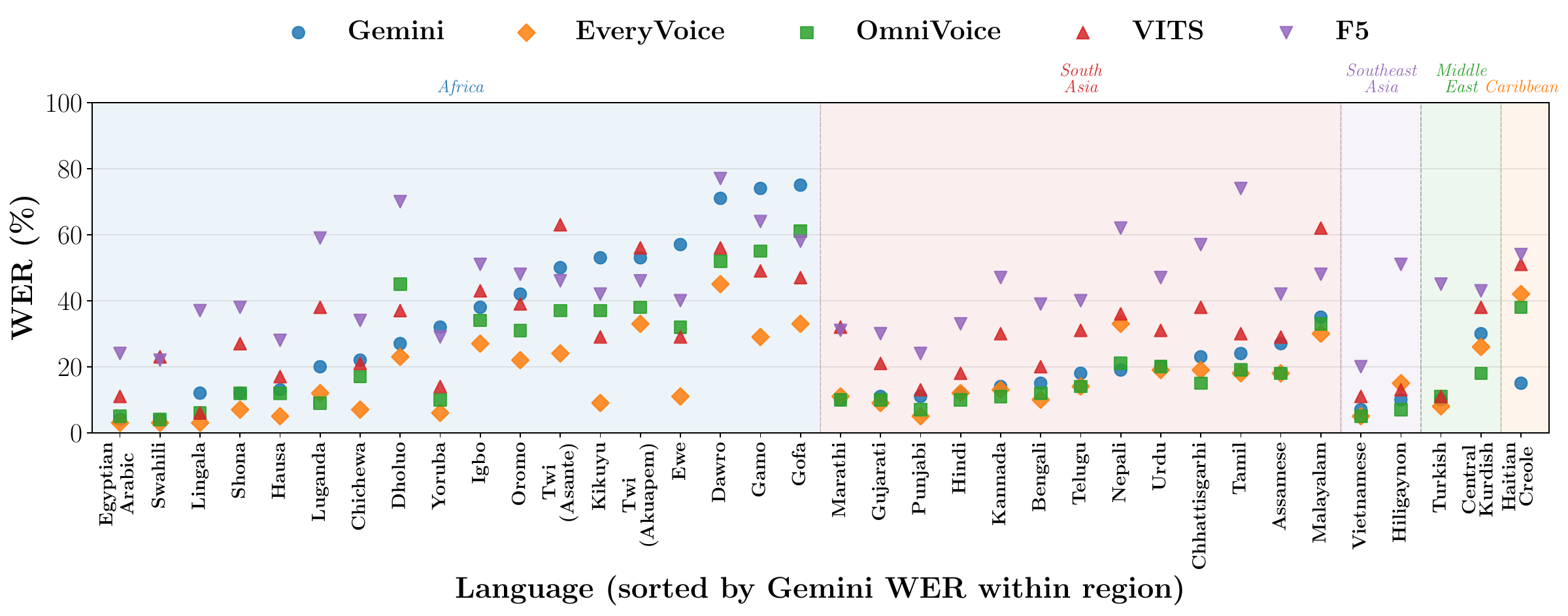}
    }
    \caption{Word error rate (WER) on the Open Bible test split for all the languages. Each bar is the mean WER over the first 500 held-out utterances per (system, language), with transcripts from \texttt{omniASR\_LLM\_1B\_v2} (lower is better).}
    \label{fig:WER_by_language}
\end{figure*}

\subsection{Main Results}

\begin{table*}[t]
\centering
\resizebox{\textwidth}{!}{%
\begin{tabular}{@{}c|cccccc|cccccc|cccccc|cc@{}}
\toprule
\multirow{2}{*}{\textbf{Language}} & \multicolumn{6}{c|}{\textbf{WER (\%) $\downarrow$}} & \multicolumn{6}{c|}{\textbf{UTMOS $\uparrow$}} & \multicolumn{6}{c|}{\textbf{MOS $\uparrow$}} & \multirow{2}{*}{$\rho_{\textbf{WER}}$} & \multirow{2}{*}{$\rho_{\textbf{UTMOS}}$} \\ \cmidrule(l){2-19}
 & \textbf{Ground truth} & \textbf{Gemini} & \textbf{OmniVoice} & \textbf{EveryVoice} & \textbf{VITS} & \textbf{F5} & \textbf{Ground truth} & \textbf{Gemini} & \textbf{OmniVoice} & \textbf{EveryVoice} & \textbf{VITS} & \textbf{F5} & \textbf{Ground truth} & \textbf{Gemini} & \textbf{OmniVoice} & \textbf{EveryVoice} & \textbf{VITS} & \textbf{F5} &  &  \\ \cmidrule(r){1-21}
Haitian Creole & \cellcolor[RGB]{200,223,241}50.52 & \cellcolor[RGB]{33,113,181}15.10 & \cellcolor[RGB]{128,185,220}38.36 & \cellcolor[RGB]{152,198,227}42.33 & \cellcolor[RGB]{204,226,242}51.31 & \cellcolor[RGB]{222,235,247}54.32 & \cellcolor[RGB]{67,162,89}3.22 & \cellcolor[RGB]{35,139,69}3.36 & \cellcolor[RGB]{210,237,206}2.72 & \cellcolor[RGB]{95,181,105}3.10 & \cellcolor[RGB]{171,220,169}2.84 & \cellcolor[RGB]{229,245,224}2.66 & \cellcolor[RGB]{184,225,182}2.70 & \cellcolor[RGB]{35,139,69}4.82 & \cellcolor[RGB]{228,245,223}2.18 & \cellcolor[RGB]{167,218,166}2.90 & \cellcolor[RGB]{212,238,208}2.37 & \cellcolor[RGB]{229,245,224}2.17 & -0.66 & \textbf{0.94} \\
Hausa & \cellcolor[RGB]{33,113,181}4.06 & \cellcolor[RGB]{87,158,205}12.71 & \cellcolor[RGB]{81,153,202}11.72 & \cellcolor[RGB]{36,116,183}4.61 & \cellcolor[RGB]{117,180,217}16.92 & \cellcolor[RGB]{222,235,247}27.64 & \cellcolor[RGB]{82,172,97}3.12 & \cellcolor[RGB]{35,139,69}3.47 & \cellcolor[RGB]{109,191,114}2.92 & \cellcolor[RGB]{134,204,135}2.77 & \cellcolor[RGB]{229,245,224}2.26 & \cellcolor[RGB]{199,232,196}2.42 & \cellcolor[RGB]{35,139,69}4.50 & \cellcolor[RGB]{71,165,91}4.13 & \cellcolor[RGB]{214,238,210}2.96 & \cellcolor[RGB]{112,193,115}3.72 & \cellcolor[RGB]{177,222,175}3.23 & \cellcolor[RGB]{229,245,224}2.85 & \textbf{-0.71} & 0.66 \\
Hindi & \cellcolor[RGB]{54,131,190}13.22 & \cellcolor[RGB]{44,122,186}11.57 & \cellcolor[RGB]{33,113,181}9.94 & \cellcolor[RGB]{49,126,188}12.38 & \cellcolor[RGB]{87,157,205}18.22 & \cellcolor[RGB]{222,235,247}32.78 & \cellcolor[RGB]{111,193,115}3.13 & \cellcolor[RGB]{35,139,69}3.52 & \cellcolor[RGB]{123,199,124}3.08 & \cellcolor[RGB]{183,225,181}2.86 & \cellcolor[RGB]{229,245,224}2.69 & \cellcolor[RGB]{99,184,108}3.19 & \cellcolor[RGB]{157,214,156}3.10 & \cellcolor[RGB]{35,139,69}4.83 & \cellcolor[RGB]{112,193,116}3.62 & \cellcolor[RGB]{128,201,129}3.43 & \cellcolor[RGB]{159,215,158}3.08 & \cellcolor[RGB]{229,245,224}2.29 & \textbf{-0.94} & 0.26 \\
Oromo & \cellcolor[RGB]{45,123,186}24.05 & \cellcolor[RGB]{166,205,231}41.83 & \cellcolor[RGB]{81,153,202}30.51 & \cellcolor[RGB]{33,113,181}21.94 & \cellcolor[RGB]{141,192,224}39.00 & \cellcolor[RGB]{222,235,247}48.30 & \cellcolor[RGB]{129,201,130}2.91 & \cellcolor[RGB]{35,139,69}3.36 & \cellcolor[RGB]{126,200,127}2.92 & \cellcolor[RGB]{187,227,185}2.70 & \cellcolor[RGB]{229,245,224}2.55 & \cellcolor[RGB]{126,200,127}2.92 & \cellcolor[RGB]{93,180,104}3.58 & \cellcolor[RGB]{229,245,224}2.50 & \cellcolor[RGB]{102,186,109}3.49 & \cellcolor[RGB]{35,139,69}4.18 & \cellcolor[RGB]{133,204,134}3.21 & \cellcolor[RGB]{147,209,147}3.11 & \textbf{-0.94} & -0.61 \\
Shona & \cellcolor[RGB]{80,152,202}16.63 & \cellcolor[RGB]{57,133,192}11.81 & \cellcolor[RGB]{56,132,191}11.50 & \cellcolor[RGB]{33,113,181}6.65 & \cellcolor[RGB]{143,193,224}27.17 & \cellcolor[RGB]{222,235,247}37.99 & \cellcolor[RGB]{229,245,224}2.65 & \cellcolor[RGB]{35,139,69}3.37 & \cellcolor[RGB]{62,158,85}3.25 & \cellcolor[RGB]{191,229,189}2.77 & \cellcolor[RGB]{213,238,209}2.70 & \cellcolor[RGB]{216,240,212}2.69 & \cellcolor[RGB]{35,139,69}4.26 & \cellcolor[RGB]{189,228,186}2.92 & \cellcolor[RGB]{212,238,208}2.75 & \cellcolor[RGB]{160,215,159}3.13 & \cellcolor[RGB]{225,243,220}2.66 & \cellcolor[RGB]{229,245,224}2.63 & -0.60 & -0.03 \\
Swahili & \cellcolor[RGB]{67,141,196}7.28 & \cellcolor[RGB]{41,119,184}3.67 & \cellcolor[RGB]{43,121,186}4.03 & \cellcolor[RGB]{33,113,181}2.64 & \cellcolor[RGB]{222,235,247}22.60 & \cellcolor[RGB]{220,234,247}22.45 & \cellcolor[RGB]{49,149,78}3.46 & \cellcolor[RGB]{35,139,69}3.54 & \cellcolor[RGB]{104,188,111}3.15 & \cellcolor[RGB]{90,178,102}3.23 & \cellcolor[RGB]{229,245,224}2.63 & \cellcolor[RGB]{147,209,147}2.96 & \cellcolor[RGB]{114,195,117}4.14 & \cellcolor[RGB]{35,139,69}4.88 & \cellcolor[RGB]{53,152,80}4.71 & \cellcolor[RGB]{70,164,90}4.55 & \cellcolor[RGB]{164,217,163}3.80 & \cellcolor[RGB]{229,245,224}3.36 & \textbf{-0.77} & \textbf{0.71} \\
Telugu & \cellcolor[RGB]{59,134,192}18.52 & \cellcolor[RGB]{53,130,190}17.55 & \cellcolor[RGB]{33,113,181}13.91 & \cellcolor[RGB]{35,115,182}14.26 & \cellcolor[RGB]{140,192,224}31.02 & \cellcolor[RGB]{222,235,247}40.44 & \cellcolor[RGB]{69,163,90}3.22 & \cellcolor[RGB]{35,139,69}3.42 & \cellcolor[RGB]{98,183,107}3.05 & \cellcolor[RGB]{120,198,121}2.93 & \cellcolor[RGB]{229,245,224}2.47 & \cellcolor[RGB]{96,182,106}3.06 & \cellcolor[RGB]{94,180,104}4.26 & \cellcolor[RGB]{35,139,69}4.77 & \cellcolor[RGB]{56,154,82}4.59 & \cellcolor[RGB]{96,182,106}4.24 & \cellcolor[RGB]{221,242,216}3.41 & \cellcolor[RGB]{229,245,224}3.36 & \textbf{-0.71} & 0.54 \\
Turkish & \cellcolor[RGB]{41,120,185}9.75 & \cellcolor[RGB]{40,119,184}9.55 & \cellcolor[RGB]{47,124,187}11.19 & \cellcolor[RGB]{33,113,181}7.76 & \cellcolor[RGB]{47,125,187}11.28 & \cellcolor[RGB]{222,235,247}45.06 & \cellcolor[RGB]{93,180,104}3.34 & \cellcolor[RGB]{35,139,69}3.50 & \cellcolor[RGB]{85,174,99}3.36 & \cellcolor[RGB]{119,197,120}3.27 & \cellcolor[RGB]{229,245,224}3.05 & \cellcolor[RGB]{75,167,93}3.39 & \cellcolor[RGB]{103,187,110}3.81 & \cellcolor[RGB]{35,139,69}4.81 & \cellcolor[RGB]{151,211,151}3.26 & \cellcolor[RGB]{144,208,144}3.33 & \cellcolor[RGB]{160,215,160}3.16 & \cellcolor[RGB]{229,245,224}2.44 & \textbf{-0.83} & 0.26 \\
Vietnamese & \cellcolor[RGB]{36,116,183}5.36 & \cellcolor[RGB]{56,132,191}7.37 & \cellcolor[RGB]{34,114,182}5.14 & \cellcolor[RGB]{33,113,181}5.00 & \cellcolor[RGB]{88,158,205}10.74 & \cellcolor[RGB]{222,235,247}20.49 & \cellcolor[RGB]{37,141,70}3.20 & \cellcolor[RGB]{35,139,69}3.21 & \cellcolor[RGB]{108,190,113}2.89 & \cellcolor[RGB]{146,209,146}2.76 & \cellcolor[RGB]{229,245,224}2.50 & \cellcolor[RGB]{159,215,158}2.72 & \cellcolor[RGB]{74,166,92}4.15 & \cellcolor[RGB]{35,139,69}4.54 & \cellcolor[RGB]{53,152,80}4.36 & \cellcolor[RGB]{139,206,139}3.56 & \cellcolor[RGB]{201,233,198}3.11 & \cellcolor[RGB]{229,245,224}2.91 & -0.49 & \textbf{0.89} \\
Yoruba & \cellcolor[RGB]{37,116,183}6.61 & \cellcolor[RGB]{222,235,247}31.93 & \cellcolor[RGB]{58,134,192}10.33 & \cellcolor[RGB]{33,113,181}5.93 & \cellcolor[RGB]{77,149,201}13.70 & \cellcolor[RGB]{194,220,239}28.75 & \cellcolor[RGB]{87,176,101}3.17 & \cellcolor[RGB]{35,139,69}3.36 & \cellcolor[RGB]{160,215,159}2.95 & \cellcolor[RGB]{156,213,156}2.96 & \cellcolor[RGB]{229,245,224}2.77 & \cellcolor[RGB]{133,203,134}3.02 & \cellcolor[RGB]{35,139,69}3.94 & \cellcolor[RGB]{212,238,208}2.36 & \cellcolor[RGB]{150,211,150}2.83 & \cellcolor[RGB]{136,205,137}2.93 & \cellcolor[RGB]{179,223,177}2.61 & \cellcolor[RGB]{229,245,224}2.23 & \textbf{-0.89} & -0.09 \\
\bottomrule
\end{tabular}%
}
\caption{\textbf{WER, UTMOSv2, and Human MOS results across the 10 languages with native-speaker ratings}. WER and UTMOSv2 are means over the first 500 held-out Open~Bible test utterances per (system, language); MOS is the mean over 340 rated clips per language (320 for Haitian Creole), pooled across three annotators and all evaluation domains. Darker shading indicates a better result. The two rightmost columns report per-language Spearman $\rho$ between WER \& MOS and UTMOSv2 \& MOS; strong correlations ($|\rho| \geq 0.7$) are \textbf{bolded}.}
\label{tab:wer-utmos-mos-comparison}
\vspace{-5mm}
\end{table*}

\begin{figure*}[t]
    \centering
    \resizebox{0.88\textwidth}{!}{%
\includegraphics{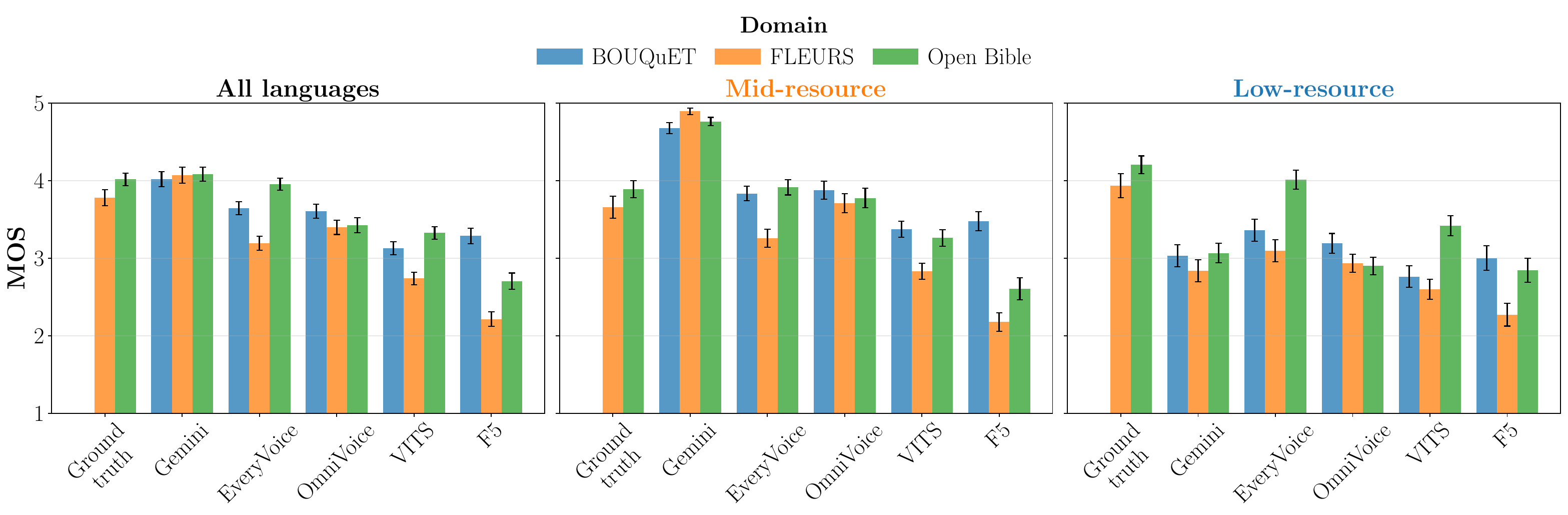}
    }
    \vspace{-2mm}
    \caption{Human MOS (mean $\pm$ 95\% CI) from the listening study, pooled over all ten languages (left) or over \textbf{Mid-resource} (Haitian Creole, Hindi, Telugu, Vietnamese, Turkish, Swahili; centre) and \textbf{Low-resource} (Hausa, Yoruba, Shona, Oromo; right) subsets. Bars group each system by \textcolor[HTML]{61B861}{Open~Bible}, \textcolor[HTML]{ff9f4a}{FLEURS}, and \textcolor[HTML]{5799c7}{BOUQuET} test domains; synthetic systems are ordered by pooled mean MOS, with ground-truth first.}
    \vspace{-2mm}
    \label{fig:MOS_by_resource_group}
\end{figure*}

\noindent \textbf{From-scratch monolingual training yields the strongest overall performance.} The results in Figure~\ref{fig:WER_by_language} demonstrate that EveryVoice typically produces the most intelligible speech with an average of 16.95\% WER across all languages, followed by OmniVoice (21.50\%) and Gemini (26.86\%), and then VITS (31.13\%) and F5 (44.51\%). We prioritize intelligibility since preserving the intended linguistic content for underrepresented languages is a prerequisite for useful TTS utilization. Under this criterion, targeted monolingual training with EveryVoice substantially outperforms larger multilingual LLM-based TTS models, even when those models have a broader cross-lingual pretraining. Specifically, while OmniVoice covers 646 languages, 
such language coverage does not reliably transfer to proximate languages.
Specifically, the large-scale language coverage of Omnivoice similarly does not guarantee strong performance on LRLs, likely due to the limited amount of low-resource language data included during pretraining.
Moreover, the performance ordering of EveryVoice, followed by VITS and then F5-TTS, mirrors the models’ increasing parameters (18.2 M, 36.3 M, 335.8 M), suggesting that more parameter-efficient architectures are better suited  
for single-speaker TTS. See Table \ref{tab:wer-utmos-by-language} for full results.

\paragraph{Model performance varies across languages within the same region.} Although the overall system ranking remains largely stable across languages, the disparities in performance between models varies dramatically within the same region. In particular, Gemini achieves strong results on supported languages but exhibits degradation on unsupported ones. Specifically, in African languages, Gemini records an average WER of 3.88\% on supported languages, which drastically increases to 40.59\% on unsupported languages within the same region, yielding a gap of 36.71\%. A similar pattern emerges in South Asia, where supported languages average 17.10\% WER, compared to 24.86\% WER for unsupported languages. 
On the other hand, OmniVoice remains comparatively stable across its supported (15.36\% WER) and unsupported (14.97\% WER) South Asian languages, probably benefiting more from related supported languages.




\paragraph{Intra-regional analysis reveals a performance gap for African languages.} African languages exhibit higher WER than languages in the South Asian, Middle Eastern, and Caribbean regional groups regardless of the model. In other words, since this pattern persists across both from-scratch and pretrained systems, it is unlikely to be explained by the limitations of a single architecture alone. Rather, the results suggest a broader mismatch between current TTS approaches and the linguistic conditions of many African languages, where tonal contrasts and orthographic variation affects synthesis quality. This gap illustrates the need for modeling strategies that are explicitly designed around the properties of African languages, rather than assuming that generic TTS architecture will transfer uniformly across different languages.

\subsection{Human Evaluation of TTS models}
\autoref{tab:wer-utmos-mos-comparison} shows the WER, UTMOS, and human MOS scores for six models across the 10 human-rated languages. Our results show that humans rated Gemini highest in six languages and EveryVoice highest in Oromo, while ground truth was preferred for three African languages (Hausa, Shona, and Yoruba). In contrast, UTMOS rated Gemini highest across all ten languages, whereas WER ranked EveryVoice best in all languages except Haitian Creole. Correlation analysis with human MOS shows that WER is strongly correlated with MOS in 7 of 10 languages ($|\rho| \geq 0.7$), while UTMOS reaches that threshold in only 3 (Haitian Creole, Swahili, Vietnamese). This suggests that WER aligns closely with human ratings than UTMOS in this setting for low-resource languages, although the variability across languages indicates that neither metric consistently captures human judgments. This result indicate that better automatic MOS-like metrics are needed for LRLs, and leave this as future work. 


\subsection{Domain Generalization}
Domain generalization capabilities of models trained on OpenBible are evaluated on the Wikipedia-based FLEURS~\citep{goyal-etal-2022-flores} and conversational BOUQuET~\citep{andrews-etal-2025-bouquet} domains. Human evaluation was conducted on 20 utterances across ten languages (see \autoref{fig:MOS_by_resource_group}).

\paragraph{Pretrained systems generalize more robustly across domains.} When considering the aggregate across all languages, the results show that Gemini achieves better generalization than models trained on the Bible. While EveryVoice, which has the highest in-domain MOS on the Bible, it shows a decline in both out-of-domain settings; on the other hand, for Omnivoice, despite a lower in-domain performance, it catches up out of domain, with VITS and F5 trailing behind. Overall, the results suggest that strong in-domain performance of an architecture does not necessarily translate to better out-of-domain generalization.


\paragraph{Domain sensitivity is language and system-dependent.} When examining MOS results by language category (mid- and low-resource), we observe that Gemini exhibits stronger domain generalization and even outperforms ground truth. Omnivoice, despite lower in-domain performance, outperforms EveryVoice in this setting. In low-resource languages, however, EveryVoice shows better generalization than the other models, including Gemini. Overall, these results suggest that domain generalization behavior is strongly influenced by both language resource level and model architecture.

\section{Conclusion}
In this paper, we have introduced \corpus{}, which is a corpus of aligned scripture speech spanning 37 underrepresented languages. Using this dataset, we have conducted a systematic comparison of five modern TTS paradigms under controlled conditions using both automatic and human evaluations across various evaluation domains. Overall, we find that no single TTS system consistently dominates across languages. EveryVoice models trained from scratch achieve the lowest mean WER of 16.95\%, which outperforms larger pretrained systems despite their broader coverage. In contrast, Gemini is preferred in six of ten human-evaluated languages because they are supported, though EveryVoice remains competitive. 
Finally, automatic metrics only partially align with human judgments; WER correlates strongly with MOS in seven languages, while UTMOS systematically favors Gemini and aligns with human evaluations only in a subset of cases. To facilitate further research, processed data, alignments, and trained models are released for all 37 languages.

\section{Limitations}


Our models are trained and evaluated primarily within a constrained, scripture-like domain characterized by narrow vocabulary, repetitive phrasing, and a relatively formal stylistic structure. Consequently, the observed gains may not directly transfer to more open-ended or conversational speech settings, where prosodic variability, discourse structure, and lexical diversity are substantially greater. 
Moreover, despite controlling for training data consistency through shared waveforms and transcripts, system-level differences in toolchain defaults—including tokenization strategies, vocoder architectures, and inference-time noise schedules—may still introduce confounding factors. These variables are not fully isolated in our experimental design and may contribute to residual performance variation across systems.

Finally, our intelligibility evaluation relies on a single ASR-based metric, namely \texttt{omniASR\_LLM\_1B\_v2}. While this provides a consistent evaluation framework across languages, it may introduce model-specific bias. In particular, Ndebele is not supported by the ASR system and is therefore excluded from WER reporting. Additionally, several low-resource languages exhibit elevated ground-truth WER, which reflects inherent recognition difficulty independent of TTS system quality.
To address this, we have reported per-language GT WER as an approximate ASR floor. Under this interpretation, cross-system WER comparisons within a given language remain meaningful for relative ranking.

\section{Ethical Considerations}

\paragraph{Data licensing.} A prerequisite for building and releasing \corpus{} was the use of speech-text data that permits redistribution and derivative reuse. All scripture audio and transcripts in our corpus are drawn from the Open Bible platform, which publishes its materials under the Creative Commons Attribution--ShareAlike (CC BY-SA) license~\citep{openbible}. Before including any language, we verified that the corresponding Open Bible collection is released under CC BY-SA, so that our aligned utterances, preprocessing scripts, and trained models can be shared with the community on compatible terms.

\paragraph{Cultural and religious content.}
Religious text can carry cultural sensitivities. We cite licensed sources, avoid implying endorsement by faith communities, and present \corpus{} strictly as a speech-technology benchmark rather than as a work of religious interpretation.

\paragraph{Human evaluation.}
Native-speaker annotators were recruited through professional platforms and direct community outreach. Ratings were collected on blinded audio clips using a disclosed 1--5 scale, personally identifying information was removed from public annotation exports, and only anonymized task metadata needed for reproducibility is released.

\section*{Acknowledgments}
This research was supported in part by the Natural Sciences and Engineering Research Council (NSERC) of Canada and in part by the AI2050 program at Schmidt Sciences. This work received partial support from Google's Gemini Academic Program Award in the form of LLM API credits. We are grateful for the support from IVADO and the Canada First Research Excellence Fund. Part of this project also received funding from the Secretaría de Ciencia, Humanidades, Tecnología e Innovación (SECIHTI).

Finally, we acknowledge the use of generative AI tools for improving proofreading and readability, no scientific content was generated
by the tools.

\bibliography{custom}

\appendix

\section{Results Per Language}
\label{sec:appendix_results_per_language}
\begin{figure}[H]
    \centering
    \includegraphics[width=\linewidth]{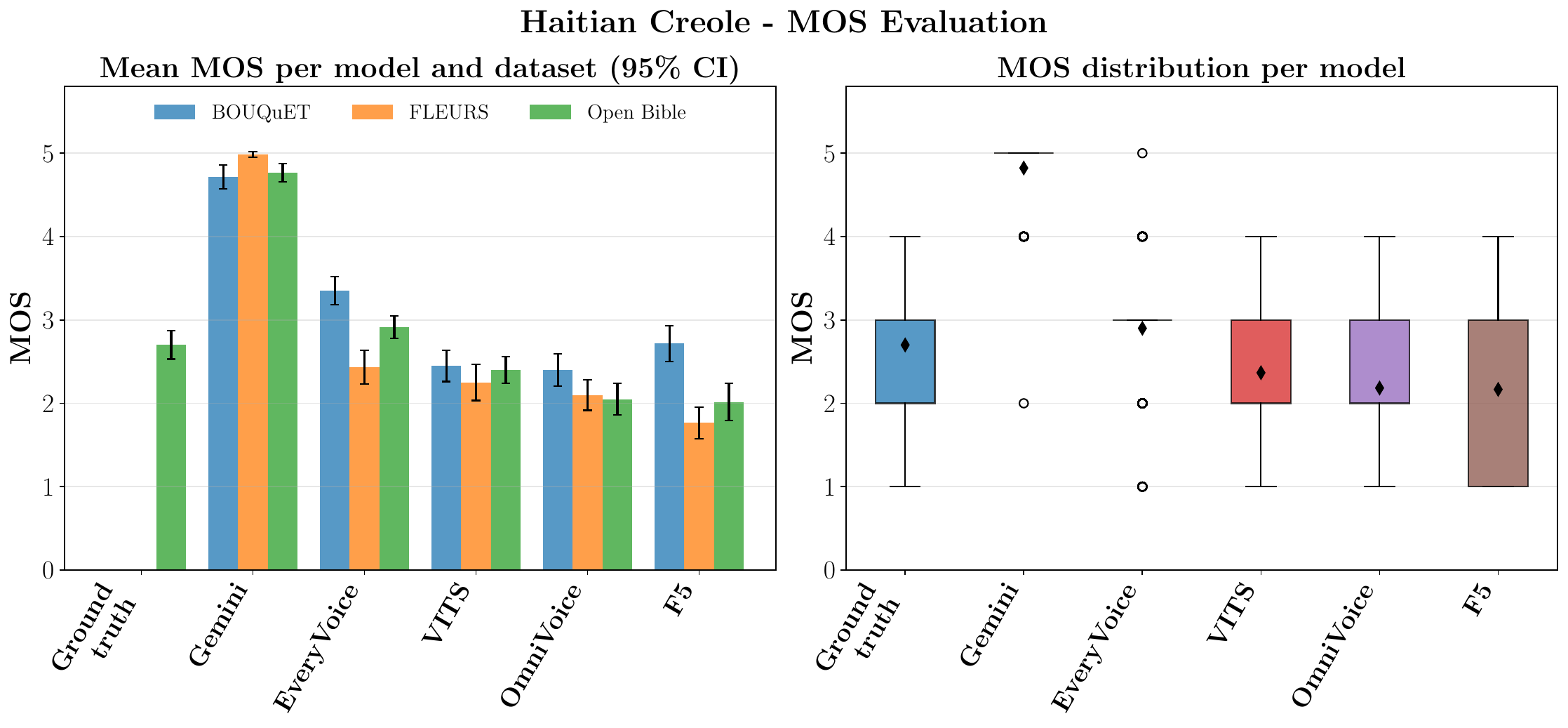}
    \caption{MOS evaluation for Haitian Creole across datasets and TTS models. Left: mean MOS with 95\% confidence intervals. Right: score distributions per model.}
    \label{fig:Haitian_Creole_MOS}
\end{figure}

\begin{figure}[H]
    \centering
    \includegraphics[width=\linewidth]{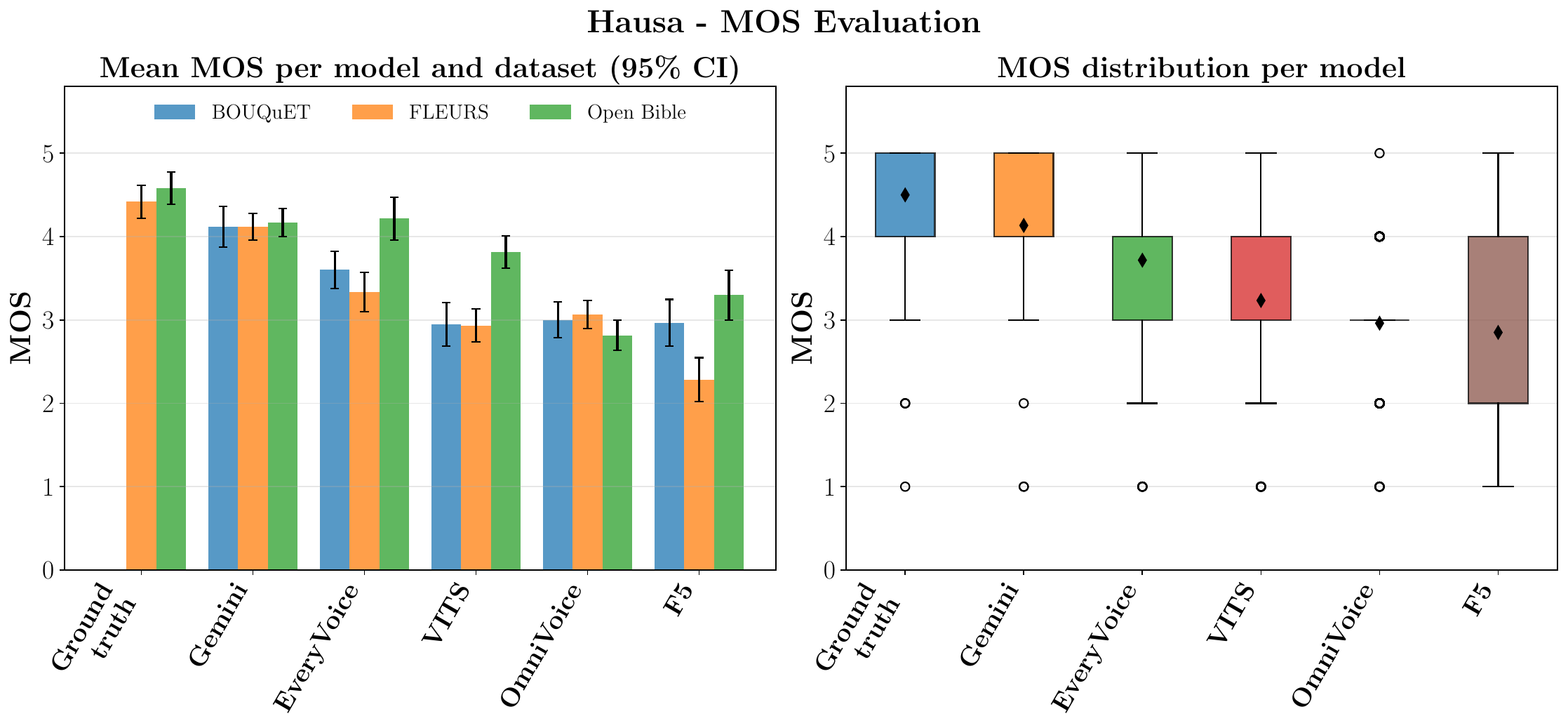}
    \caption{MOS evaluation for Hausa across datasets and TTS models. Left: mean MOS with 95\% confidence intervals. Right: score distributions per model.}
    \label{fig:Hausa_MOS}
\end{figure}

\begin{figure}[H]
    \centering
    \includegraphics[width=\linewidth]{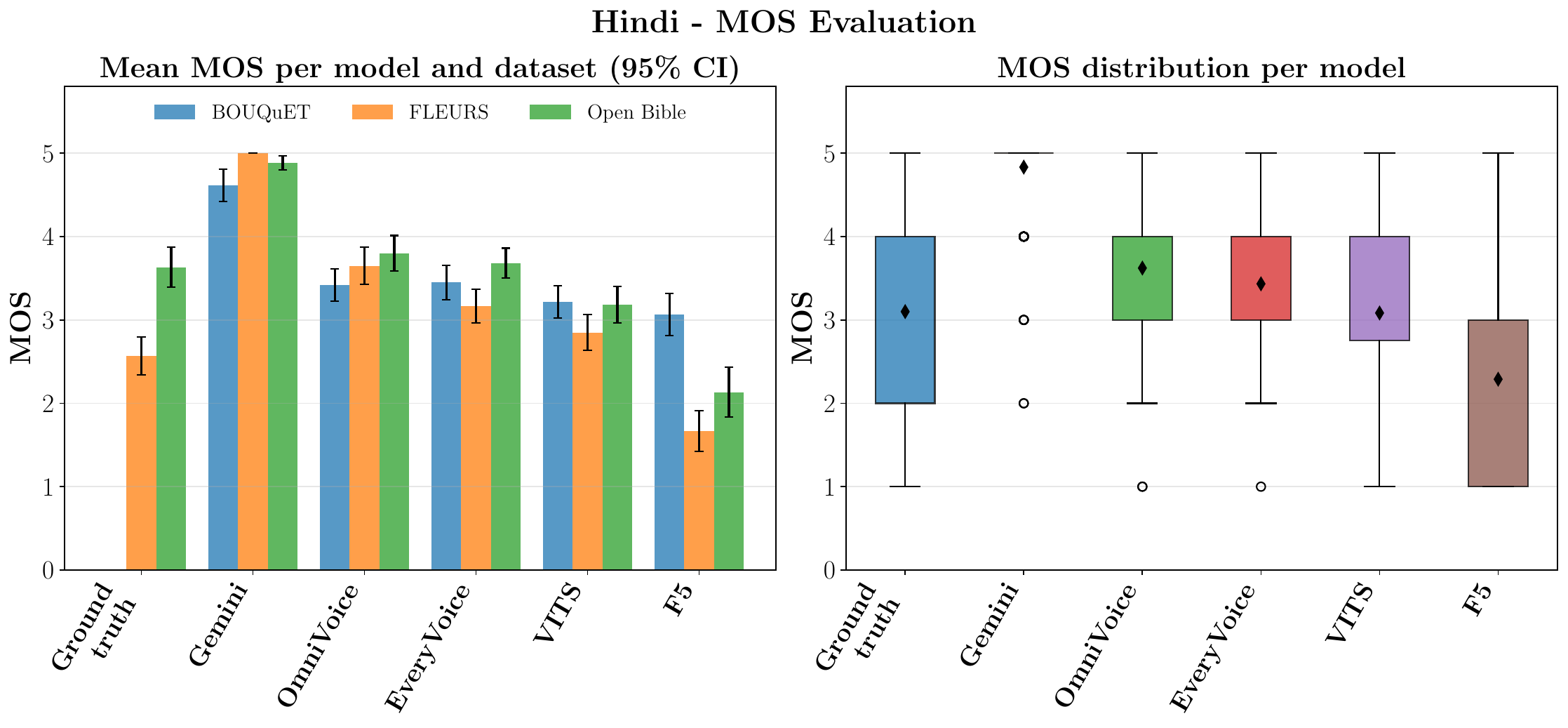}
    \caption{MOS evaluation for Hindi across datasets and TTS models. Left: mean MOS with 95\% confidence intervals. Right: score distributions per model.}
    \label{fig:Hindi_MOS}
\end{figure}

\begin{figure}[H]
    \centering
    \includegraphics[width=\linewidth]{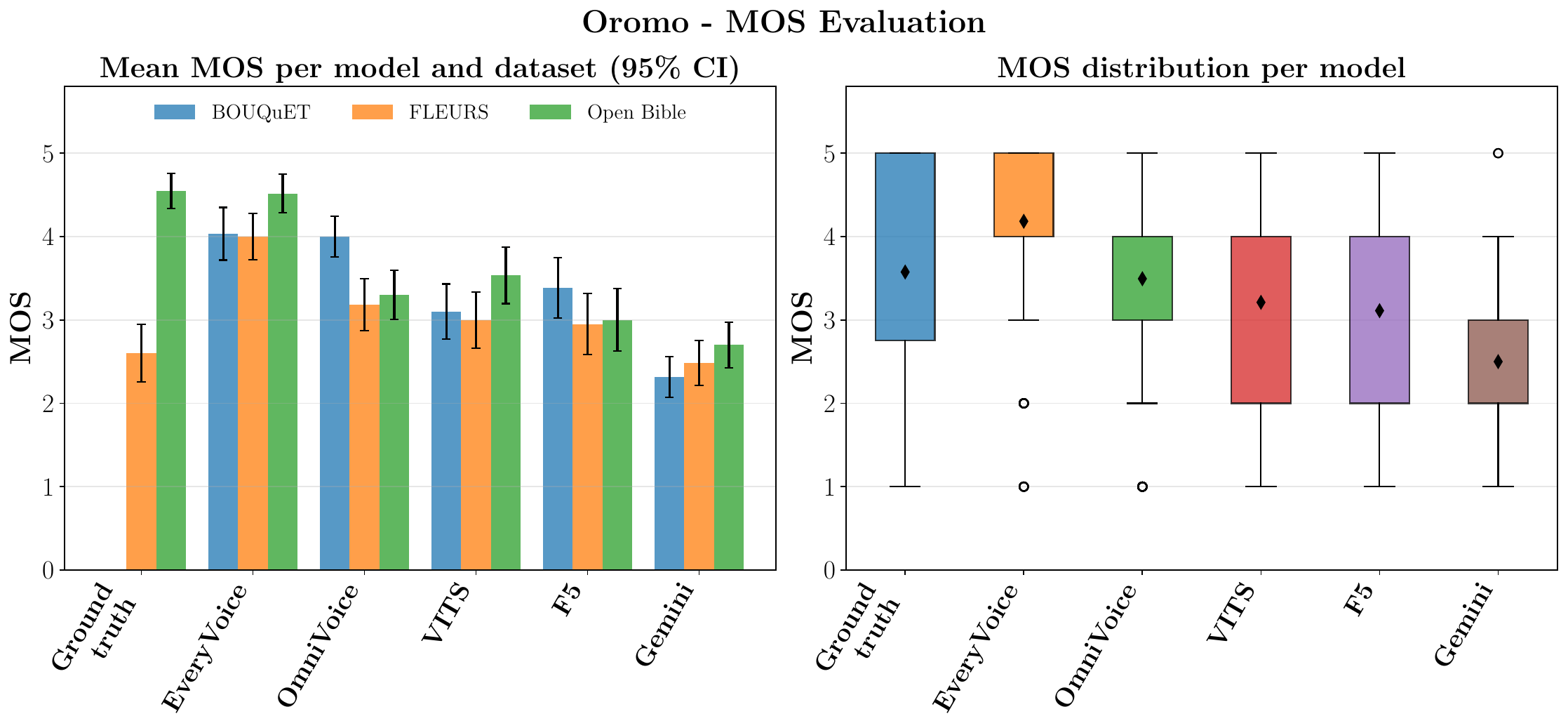}
    \caption{MOS evaluation for Oromo across datasets and TTS models. Left: mean MOS with 95\% confidence intervals. Right: score distributions per model.}
    \label{fig:Oromo_MOS}
\end{figure}

\begin{figure}[H]
    \centering
    \includegraphics[width=\linewidth]{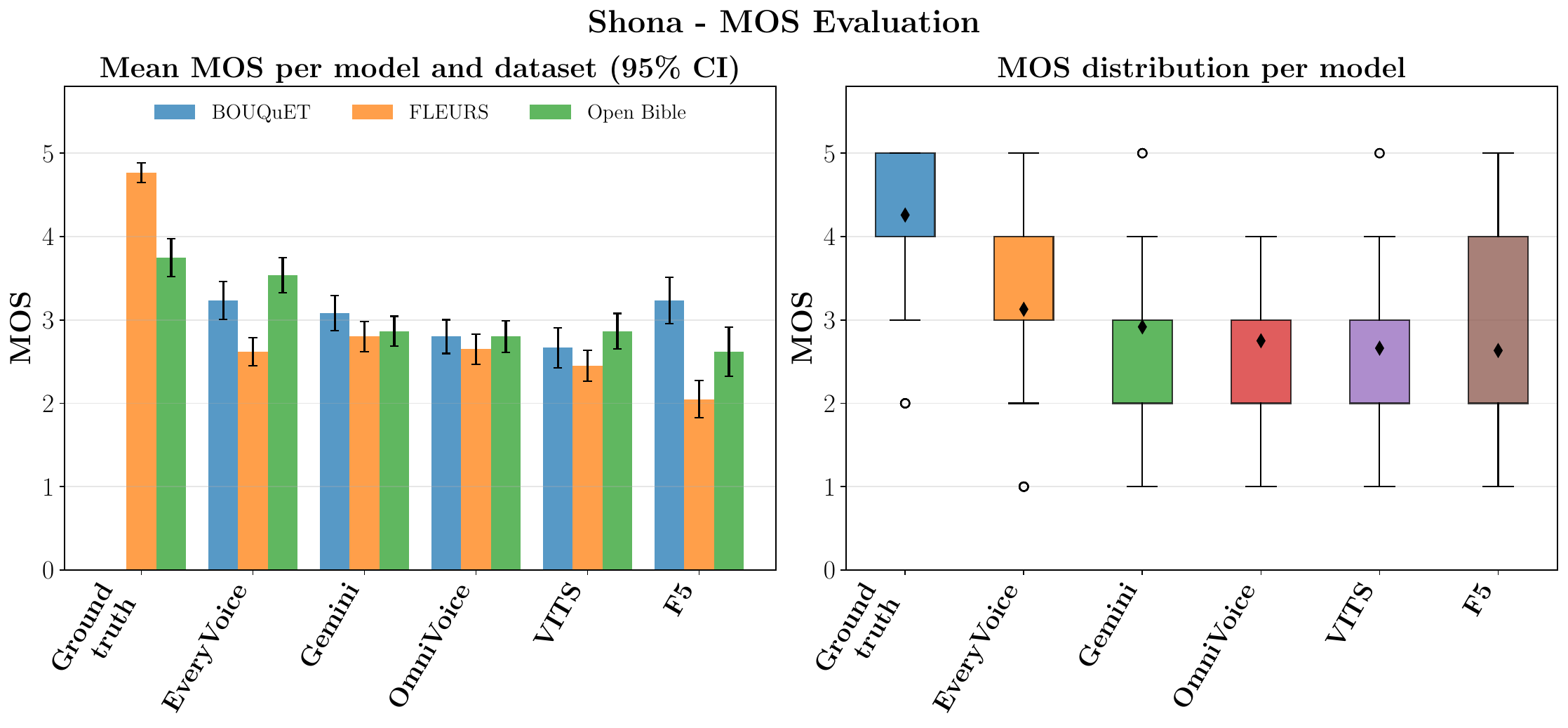}
    \caption{MOS evaluation for Shona across datasets and TTS models. Left: mean MOS with 95\% confidence intervals. Right: score distributions per model.}
    \label{fig:Shona_MOS}
\end{figure}

\begin{figure}[H]
    \centering
    \includegraphics[width=\linewidth]{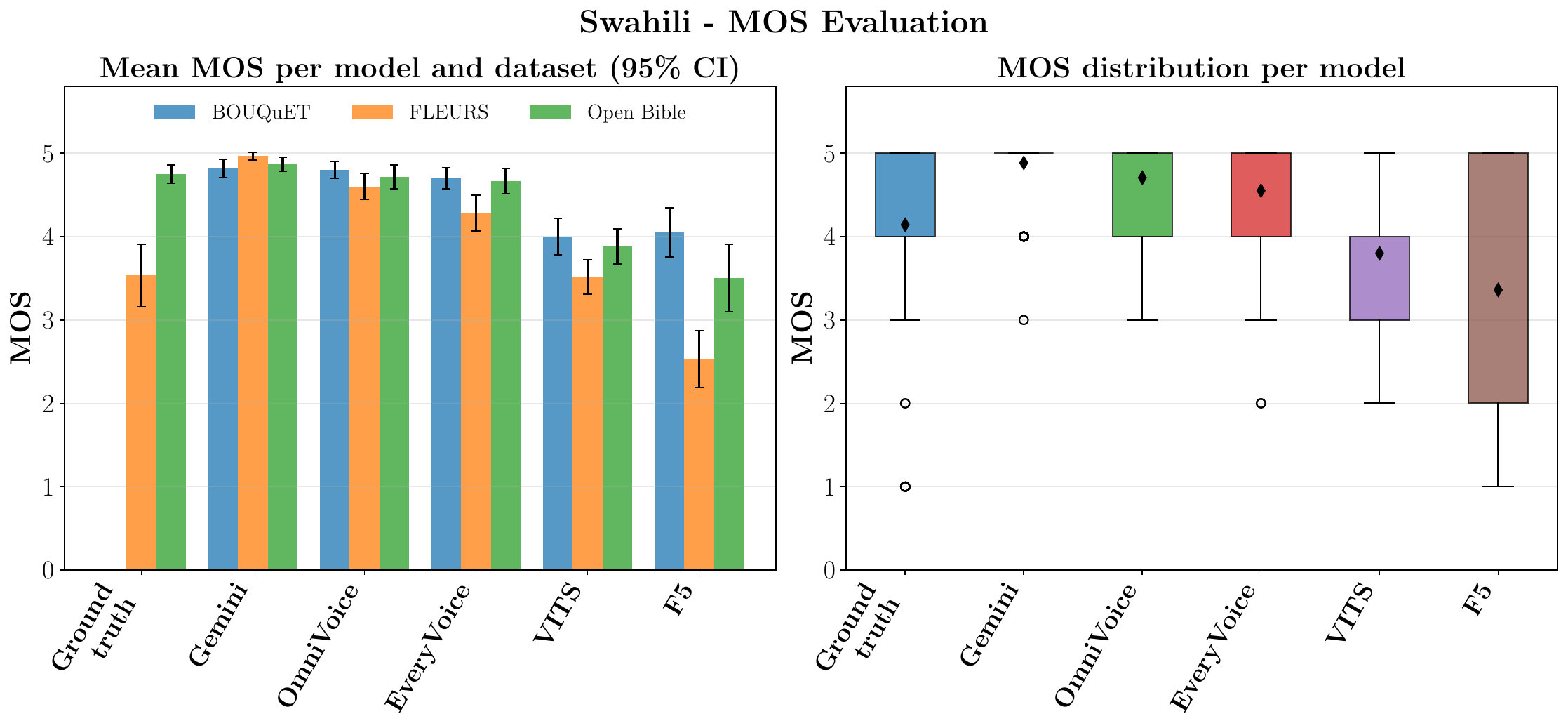}
    \caption{MOS evaluation for Swahili across datasets and TTS models. Left: mean MOS with 95\% confidence intervals. Right: score distributions per model.}
    \label{fig:Swahili_MOS}
\end{figure}

\begin{figure}[H]
    \centering
    \includegraphics[width=\linewidth]{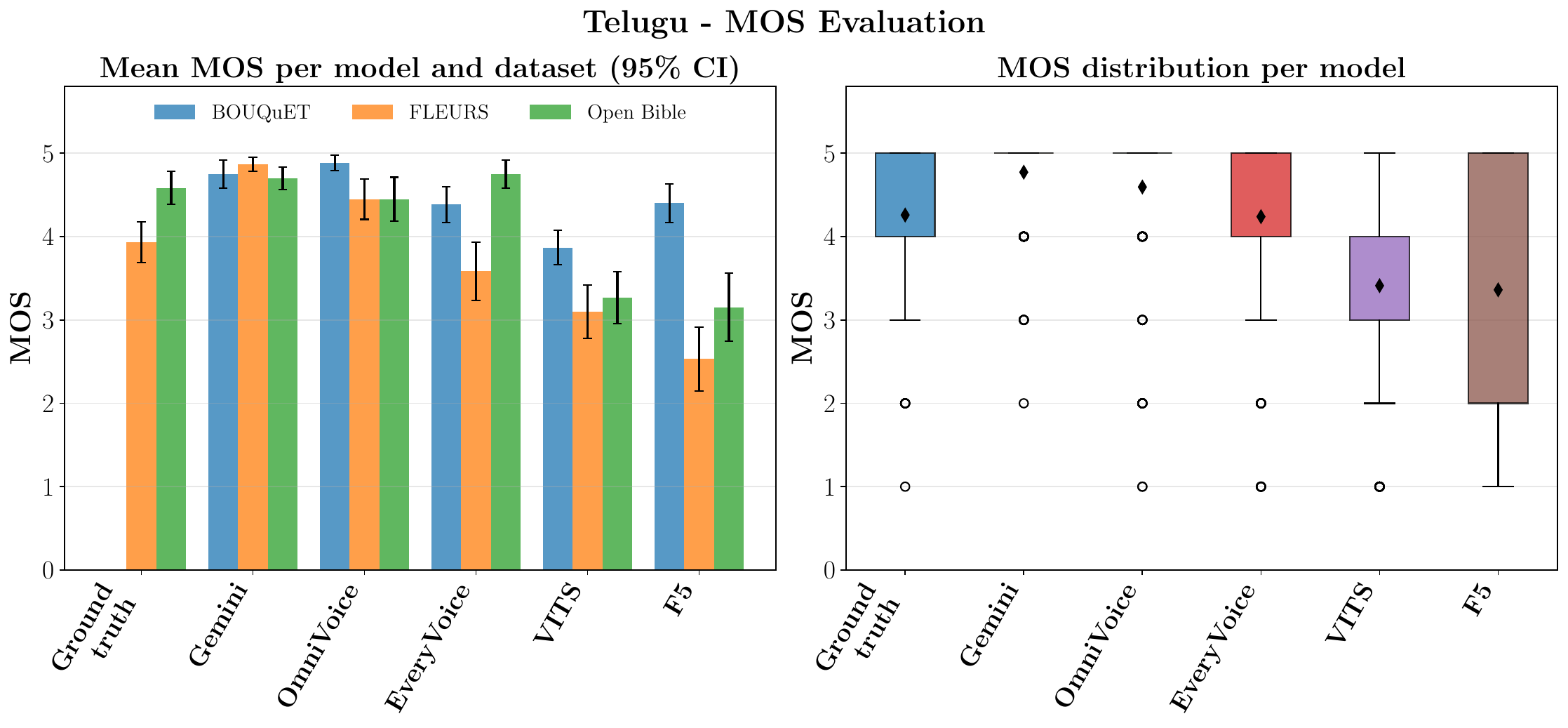}
    \caption{MOS evaluation for Telugu across datasets and TTS models. Left: mean MOS with 95\% confidence intervals. Right: score distributions per model.}
    \label{fig:Telugu_MOS}
\end{figure}

\begin{figure}[H]
    \centering
    \includegraphics[width=\linewidth]{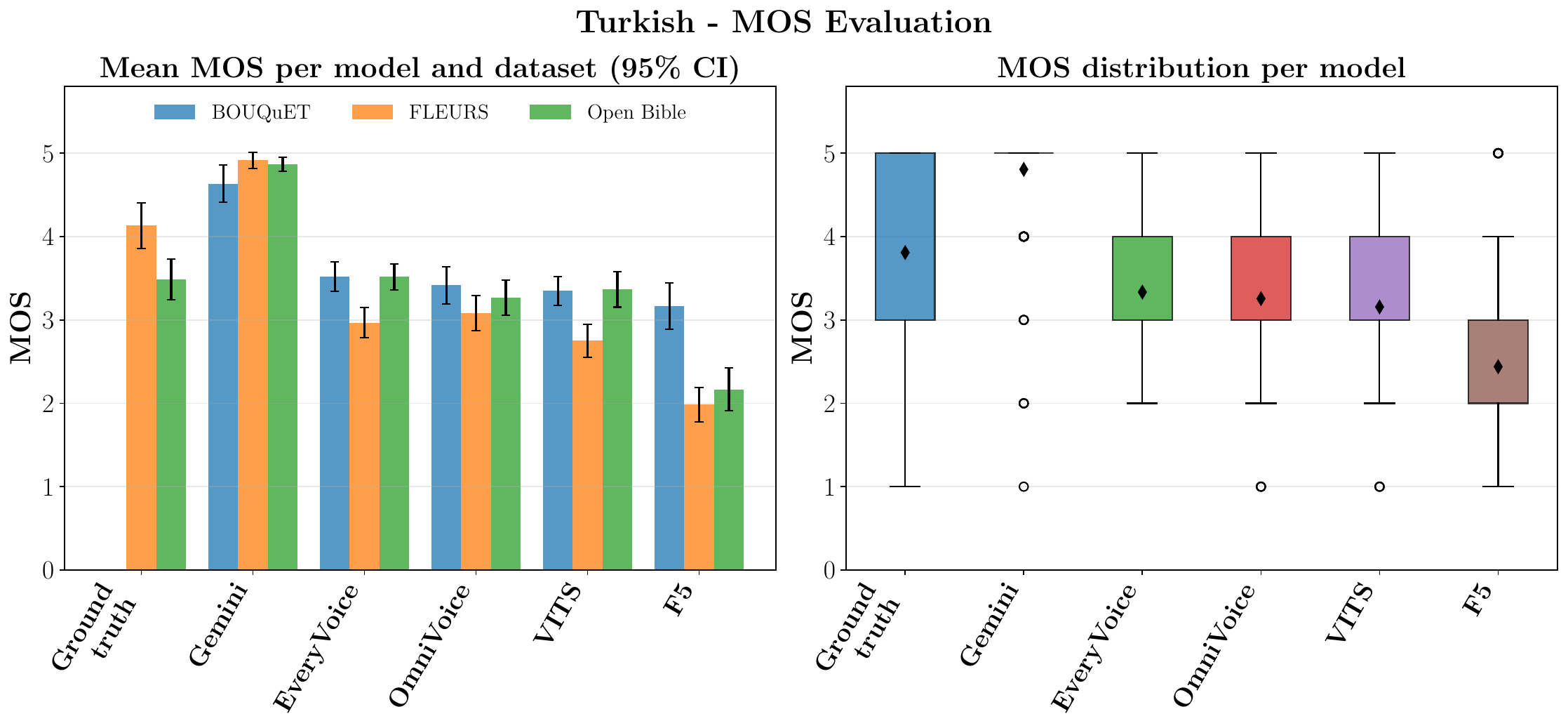}
    \caption{MOS evaluation for Turkish across datasets and TTS models. Left: mean MOS with 95\% confidence intervals. Right: score distributions per model.}
    \label{fig:Turkish_MOS}
\end{figure}

\begin{figure}[H]
    \centering
    \includegraphics[width=\linewidth]{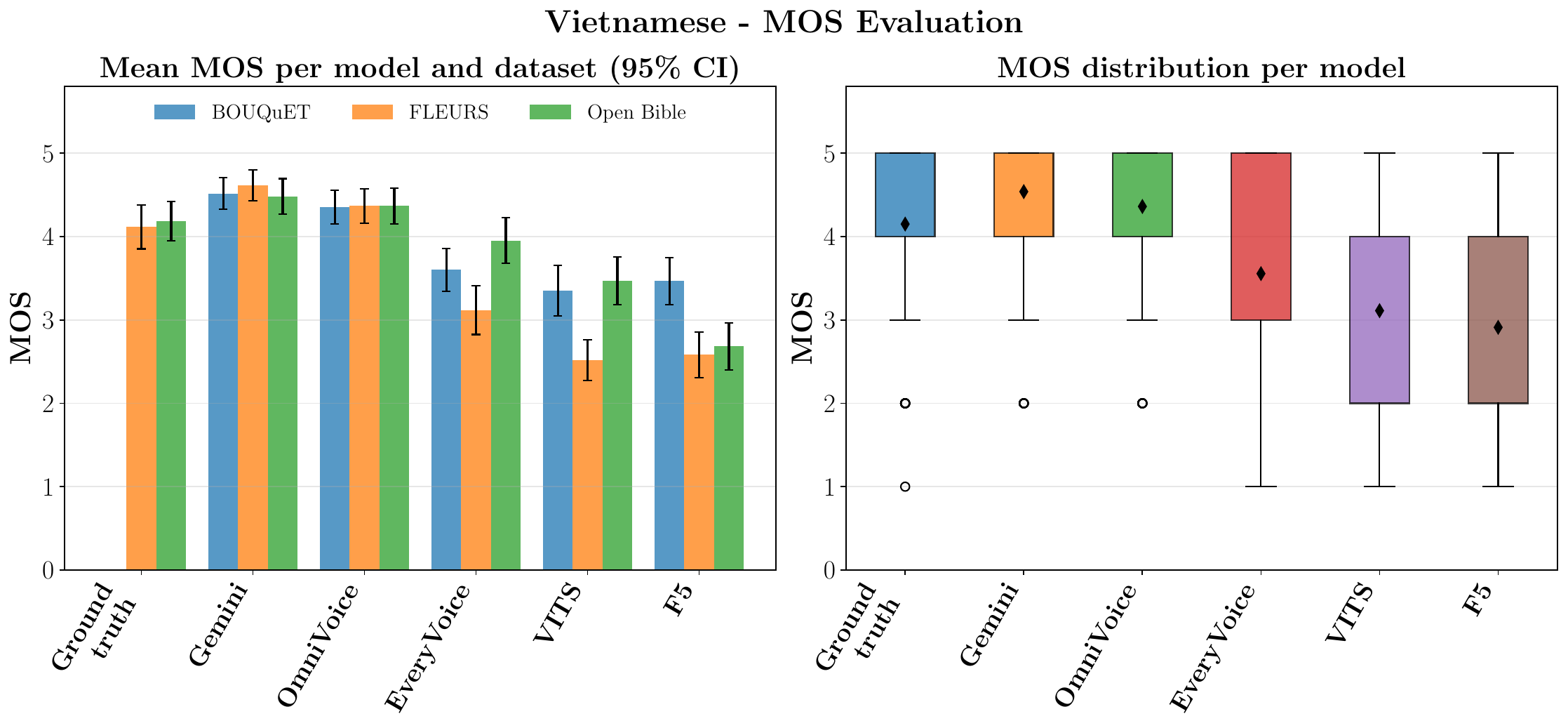}
    \caption{MOS evaluation for Vietnamese across datasets and TTS models. Left: mean MOS with 95\% confidence intervals. Right: score distributions per model.}
    \label{fig:Vietnamese_MOS}
\end{figure}

\begin{figure}[H]
    \centering
    \includegraphics[width=\linewidth]{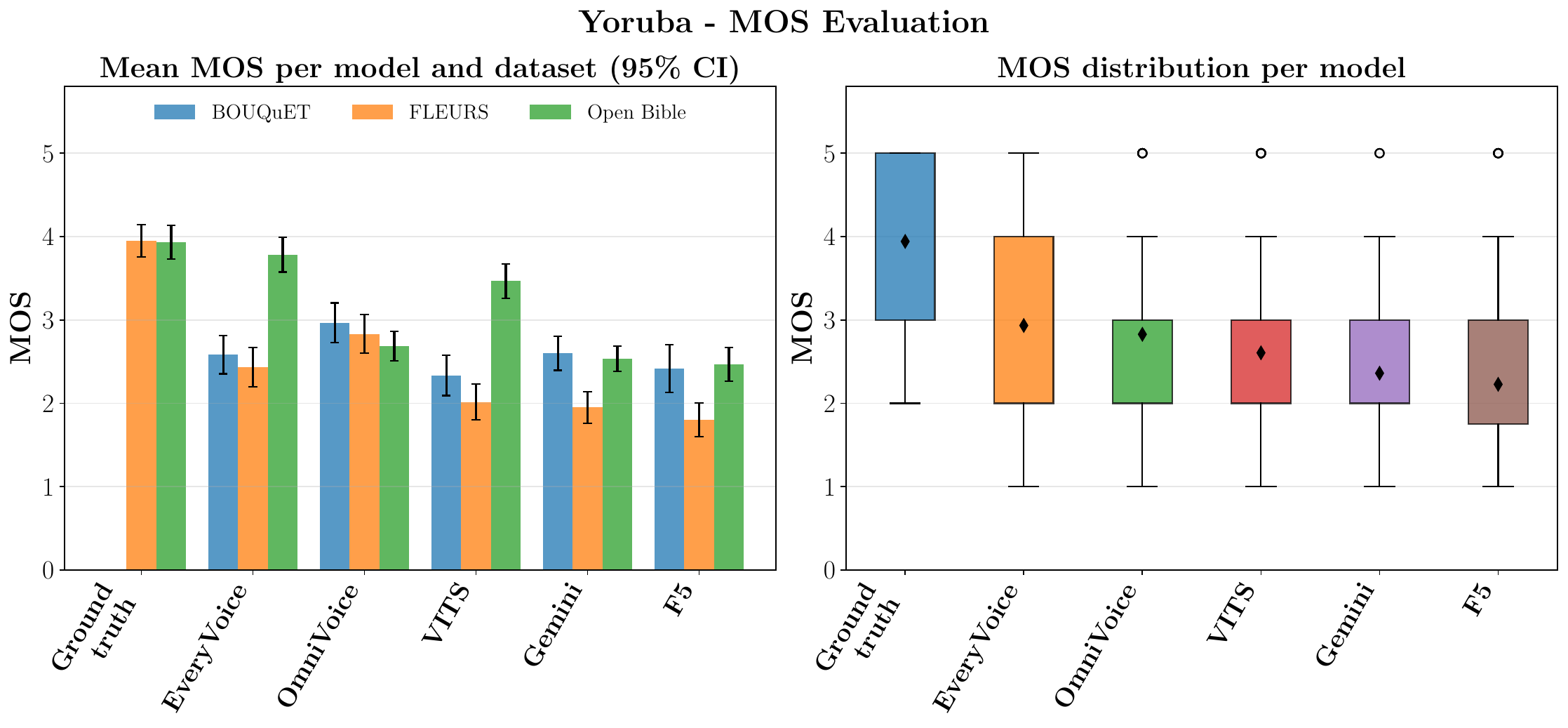}
    \caption{MOS evaluation for Yoruba across datasets and TTS models. Left: mean MOS with 95\% confidence intervals. Right: score distributions per model.}
    \label{fig:Yoruba_MOS}
\end{figure}

Table~\ref{tab:wer-utmos-by-language} reports the full per-language WER and UTMOSv2 scores for all five TTS systems and ground truth across all 37 \corpus{} languages, computed on the first 500 held-out utterances of the Open Bible test split. Figures \ref{fig:Haitian_Creole_MOS}--\ref{fig:Yoruba_MOS} show the per-language human MOS results for each of the 10 languages in the listening study, broken down by evaluation domain (Open~Bible, FLEURS, BOUQuET). Each figure displays mean MOS with 95\% confidence intervals alongside score distributions per model, allowing a detailed view of system behaviour and annotator agreement within each language.

\begin{table*}[h]
\centering
\resizebox{\textwidth}{!}{%
\begin{tabular}{@{}c|cc|cc|cc|cc|cc|cc@{}}
\toprule
\multirow{2}{*}{\textbf{Language}} & \multicolumn{2}{c|}{\textbf{Ground truth}} & \multicolumn{2}{c|}{\textbf{Gemini}} & \multicolumn{2}{c|}{\textbf{EveryVoice}} & \multicolumn{2}{c|}{\textbf{OmniVoice}} & \multicolumn{2}{c|}{\textbf{VITS}} & \multicolumn{2}{c}{\textbf{F5}} \\ \cmidrule(l){2-13}
 & \textbf{WER (\%) $\downarrow$} & \textbf{UTMOS $\uparrow$} & \textbf{WER (\%) $\downarrow$} & \textbf{UTMOS $\uparrow$} & \textbf{WER (\%) $\downarrow$} & \textbf{UTMOS $\uparrow$} & \textbf{WER (\%) $\downarrow$} & \textbf{UTMOS $\uparrow$} & \textbf{WER (\%) $\downarrow$} & \textbf{UTMOS $\uparrow$} & \textbf{WER (\%) $\downarrow$} & \textbf{UTMOS $\uparrow$} \\ \cmidrule(r){1-13}
Egyptian Arabic & 5.28 ± 10.35 & 3.02 ± 0.28 & 4.08 ± 9.79 & \textbf{3.40 ± 0.29} & \textbf{2.91 ± 6.77} & 2.75 ± 0.26 & 4.61 ± 7.87 & 3.02 ± 0.24 & 11.47 ± 11.26 & 2.77 ± 0.29 & 24.47 ± 27.39 & 2.94 ± 0.3 \\
Assamese & 21.03 ± 15.24 & 2.67 ± 0.35 & 26.52 ± 16.33 & \textbf{3.55 ± 0.27} & 18.40 ± 14.95 & 2.48 ± 0.31 & \textbf{18.31 ± 14.36} & 3.06 ± 0.26 & 28.60 ± 15.72 & 2.23 ± 0.29 & 41.70 ± 21.93 & 2.88 ± 0.31 \\
Bengali & 12.25 ± 11.29 & 2.91 ± 0.34 & 15.11 ± 12.36 & \textbf{3.47 ± 0.28} & \textbf{10.49 ± 10.56} & 2.76 ± 0.32 & 11.88 ± 11.45 & 3.03 ± 0.25 & 19.58 ± 13.25 & 2.29 ± 0.29 & 38.90 ± 21.28 & 2.93 ± 0.3 \\
Central Kurdish & 23.58 ± 17.75 & 3.33 ± 0.26 & 30.12 ± 18.33 & \textbf{3.47 ± 0.28} & 26.15 ± 19.77 & 3.08 ± 0.28 & \textbf{18.07 ± 14.96} & 3.00 ± 0.24 & 37.56 ± 18.06 & 2.66 ± 0.29 & 43.15 ± 25.19 & 3.13 ± 0.24 \\
Chhattisgarhi & 26.21 ± 23.74 & 3.28 ± 0.28 & 23.19 ± 13.27 & \textbf{3.46 ± 0.30} & 18.72 ± 16.41 & 3.02 ± 0.26 & \textbf{14.97 ± 11.06} & 2.86 ± 0.30 & 37.93 ± 24.05 & 2.79 ± 0.29 & 56.89 ± 21.32 & 3.06 ± 0.25 \\
Chichewa & 12.13 ± 47.38 & 2.67 ± 0.39 & 21.60 ± 14.27 & \textbf{3.48 ± 0.29} & \textbf{7.27 ± 8.86} & 2.72 ± 0.32 & 16.80 ± 13.59 & 2.81 ± 0.34 & 21.19 ± 12.99 & 2.55 ± 0.38 & 34.04 ± 27.07 & 2.48 ± 0.34 \\
Dawro & 49.32 ± 16.82 & 2.88 ± 0.36 & 71.15 ± 15.59 & \textbf{3.34 ± 0.31} & \textbf{45.15 ± 16.05} & 2.55 ± 0.31 & 52.34 ± 16.35 & 3.19 ± 0.25 & 56.18 ± 17.20 & 2.38 ± 0.38 & 77.06 ± 15.17 & 2.9 ± 0.28 \\
Dholuo & 24.37 ± 16.31 & 3.32 ± 0.31 & 27.27 ± 17.94 & \textbf{3.38 ± 0.33} & \textbf{22.63 ± 16.77} & 3.06 ± 0.29 & 44.80 ± 20.09 & 3.31 ± 0.24 & 37.50 ± 18.56 & 2.68 ± 0.31 & 70.28 ± 21.94 & 2.96 ± 0.32 \\
Ewe & 14.04 ± 12.22 & \textbf{3.26 ± 0.31} & 57.09 ± 19.18 & 3.24 ± 0.37 & \textbf{11.26 ± 11.79} & 3.06 ± 0.25 & 32.01 ± 17.71 & 2.93 ± 0.35 & 28.52 ± 16.42 & 2.46 ± 0.32 & 39.79 ± 20.64 & 2.94 ± 0.3 \\
Gamo & 34.82 ± 19.48 & 2.93 ± 0.35 & 73.52 ± 28.21 & \textbf{3.31 ± 0.33} & \textbf{29.43 ± 15.22} & 2.79 ± 0.25 & 55.13 ± 18.64 & 2.71 ± 0.29 & 49.27 ± 18.29 & 2.47 ± 0.32 & 64.01 ± 21.27 & 2.74 ± 0.29 \\
Gofa & 38.19 ± 18.59 & 3.15 ± 0.30 & 74.75 ± 26.79 & \textbf{3.38 ± 0.33} & \textbf{33.45 ± 14.49} & 3.09 ± 0.25 & 60.53 ± 20.22 & 3.02 ± 0.24 & 47.22 ± 18.35 & 2.77 ± 0.32 & 58.43 ± 22.64 & 2.8 ± 0.26 \\
Gujarati & 10.62 ± 11.36 & \textbf{3.65 ± 0.23} & 11.21 ± 10.71 & 3.54 ± 0.26 & \textbf{9.02 ± 9.68} & 3.39 ± 0.22 & 9.58 ± 10.40 & 3.27 ± 0.22 & 20.63 ± 39.90 & 3.32 ± 0.26 & 30.13 ± 22.97 & 3.22 ± 0.24 \\
Haitian Creole & 50.52 ± 23.18 & 3.22 ± 0.36 & \textbf{15.10 ± 12.39} & \textbf{3.36 ± 0.34} & 42.33 ± 23.80 & 3.1 ± 0.28 & 38.36 ± 18.11 & 2.72 ± 0.28 & 51.31 ± 21.24 & 2.84 ± 0.33 & 54.32 ± 22.89 & 2.66 ± 0.26 \\
Hausa & \textbf{4.06 ± 9.93} & 3.12 ± 0.37 & 12.71 ± 17.86 & \textbf{3.47 ± 0.32} & 4.61 ± 12.90 & 2.77 ± 0.29 & 11.72 ± 12.55 & 2.92 ± 0.26 & 16.92 ± 20.69 & 2.26 ± 0.33 & 27.64 ± 20.96 & 2.42 ± 0.31 \\
Hiligaynon & 8.88 ± 12.35 & 3.33 ± 0.30 & 9.85 ± 9.82 & \textbf{3.48 ± 0.30} & 15.43 ± 23.17 & 3.15 ± 0.34 & \textbf{7.08 ± 7.43} & 2.97 ± 0.29 & 13.47 ± 14.50 & 2.95 ± 0.35 & 51.18 ± 22.30 & 3.06 ± 0.36 \\
Hindi & 13.22 ± 9.88 & 3.13 ± 0.28 & 11.57 ± 10.87 & \textbf{3.52 ± 0.28} & 12.38 ± 9.57 & 2.86 ± 0.26 & \textbf{9.94 ± 9.05} & 3.08 ± 0.24 & 18.22 ± 12.35 & 2.69 ± 0.27 & 32.78 ± 17.96 & 3.19 ± 0.27 \\
Igbo & \textbf{27.12 ± 14.71} & \textbf{3.42 ± 0.29} & 37.77 ± 15.13 & 3.23 ± 0.37 & 27.15 ± 17.97 & 3.19 ± 0.29 & 33.97 ± 14.62 & 3.06 ± 0.24 & 42.96 ± 20.63 & 2.64 ± 0.37 & 51.19 ± 27.26 & 3.03 ± 0.27 \\
Kannada & 15.30 ± 15.56 & 2.94 ± 0.31 & 14.30 ± 14.99 & \textbf{3.44 ± 0.28} & 12.71 ± 14.59 & 2.74 ± 0.29 & \textbf{11.13 ± 12.99} & 3.04 ± 0.24 & 30.29 ± 16.85 & 2.4 ± 0.27 & 47.29 ± 25.12 & 2.76 ± 0.29 \\
Kikuyu & 11.61 ± 14.95 & 2.45 ± 0.32 & 52.94 ± 31.13 & \textbf{3.35 ± 0.31} & \textbf{8.51 ± 12.04} & 2.62 ± 0.28 & 36.92 ± 16.39 & 3.23 ± 0.24 & 28.65 ± 16.77 & 2.47 ± 0.31 & 41.76 ± 25.22 & 2.79 ± 0.28 \\
Lingala & 3.17 ± 6.18 & 3.19 ± 0.30 & 11.95 ± 8.54 & \textbf{3.49 ± 0.30} & \textbf{2.63 ± 5.87} & 2.98 ± 0.26 & 5.92 ± 7.34 & 3.05 ± 0.28 & 5.64 ± 6.89 & 2.8 ± 0.33 & 36.99 ± 18.72 & 2.88 ± 0.28 \\
Luganda & 13.07 ± 13.93 & 2.85 ± 0.30 & 20.16 ± 14.47 & \textbf{3.36 ± 0.31} & 11.80 ± 18.43 & 2.75 ± 0.28 & \textbf{9.49 ± 10.58} & 3.25 ± 0.24 & 38.09 ± 32.79 & 2.67 ± 0.33 & 59.50 ± 23.81 & 2.51 ± 0.3 \\
Malayalam & \textbf{26.77 ± 21.46} & 3.11 ± 0.30 & 34.87 ± 19.70 & \textbf{3.44 ± 0.30} & 29.65 ± 17.16 & 2.80 ± 0.29 & 32.69 ± 21.49 & 2.86 ± 0.30 & 62.40 ± 23.29 & 2.28 ± 0.34 & 47.71 ± 25.25 & 2.98 ± 0.27 \\
Marathi & 14.42 ± 17.47 & 2.55 ± 0.44 & \textbf{9.65 ± 11.60} & \textbf{3.43 ± 0.31} & 11.46 ± 12.70 & 2.19 ± 0.32 & 10.06 ± 19.01 & 2.94 ± 0.26 & 32.39 ± 17.11 & 1.99 ± 0.32 & 30.95 ± 25.04 & 2.42 ± 0.31 \\
Ndebele & -- & 2.85 ± 0.36 & -- & \textbf{3.39 ± 0.30} & -- & 2.58 ± 0.29 & -- & 3.04 ± 0.27 & -- & 2.53 ± 0.31 & -- & 2.81 ± 0.35 \\
Nepali & 24.95 ± 18.74 & 3.31 ± 0.27 & \textbf{19.04 ± 13.47} & \textbf{3.51 ± 0.27} & 32.83 ± 22.62 & 3.04 ± 0.28 & 20.59 ± 15.31 & 2.38 ± 0.30 & 36.48 ± 19.19 & 2.32 ± 0.34 & 61.82 ± 24.18 & 2.99 ± 0.31 \\
Oromo & 24.05 ± 18.10 & 2.91 ± 0.35 & 41.83 ± 21.57 & \textbf{3.36 ± 0.31} & \textbf{21.94 ± 19.20} & 2.7 ± 0.3 & 30.51 ± 20.46 & 2.92 ± 0.26 & 39.00 ± 23.38 & 2.55 ± 0.32 & 48.30 ± 30.77 & 2.92 ± 0.27 \\
Punjabi & 6.96 ± 8.67 & 2.94 ± 0.28 & 11.16 ± 9.33 & \textbf{3.47 ± 0.29} & \textbf{4.72 ± 7.19} & 2.67 ± 0.26 & 7.09 ± 8.23 & 2.78 ± 0.25 & 12.68 ± 12.00 & 2.6 ± 0.28 & 24.50 ± 20.69 & 2.96 ± 0.23 \\
Shona & 16.63 ± 29.00 & 2.65 ± 0.29 & 11.81 ± 11.20 & \textbf{3.37 ± 0.31} & \textbf{6.65 ± 13.83} & 2.77 ± 0.24 & 11.50 ± 13.36 & 3.25 ± 0.22 & 27.17 ± 24.07 & 2.7 ± 0.29 & 37.99 ± 35.53 & 2.69 ± 0.25 \\
Swahili & 7.28 ± 9.83 & 3.46 ± 0.30 & 3.67 ± 6.67 & \textbf{3.54 ± 0.30} & \textbf{2.64 ± 5.66} & 3.23 ± 0.22 & 4.03 ± 7.32 & 3.15 ± 0.25 & 22.60 ± 15.95 & 2.63 ± 0.29 & 22.45 ± 20.91 & 2.96 ± 0.29 \\
Tamil & 20.46 ± 18.48 & 2.96 ± 0.37 & 23.69 ± 20.78 & \textbf{3.35 ± 0.31} & \textbf{18.19 ± 19.19} & 2.93 ± 0.35 & 19.28 ± 18.38 & 3.21 ± 0.23 & 29.79 ± 19.97 & 2.62 ± 0.33 & 73.60 ± 27.17 & 3.07 ± 0.31 \\
Telugu & 18.52 ± 17.20 & 3.22 ± 0.28 & 17.55 ± 14.44 & \textbf{3.42 ± 0.30} & 14.26 ± 14.37 & 2.93 ± 0.26 & \textbf{13.91 ± 13.50} & 3.05 ± 0.28 & 31.02 ± 16.65 & 2.47 ± 0.31 & 40.44 ± 27.47 & 3.06 ± 0.24 \\
Turkish & 9.75 ± 13.43 & 3.34 ± 0.21 & 9.55 ± 13.88 & \textbf{3.50 ± 0.29} & \textbf{7.76 ± 12.18} & 3.27 ± 0.22 & 11.19 ± 12.71 & 3.36 ± 0.23 & 11.28 ± 13.45 & 3.05 ± 0.29 & 45.06 ± 25.19 & 3.39 ± 0.27 \\
Twi (Akuapem) & \textbf{27.56 ± 18.33} & 3.30 ± 0.31 & 52.69 ± 18.91 & \textbf{3.36 ± 0.35} & 32.93 ± 17.57 & 3.21 ± 0.27 & 36.92 ± 19.89 & 3.10 ± 0.27 & 56.25 ± 23.35 & 2.4 ± 0.33 & 46.10 ± 23.13 & 2.99 ± 0.35 \\
Twi (Asante) & 27.19 ± 15.15 & \textbf{3.50 ± 0.31} & 50.23 ± 18.95 & 3.34 ± 0.36 & \textbf{24.35 ± 16.91} & 3.28 ± 0.29 & 37.44 ± 16.97 & 2.92 ± 0.27 & 62.69 ± 21.23 & 2.4 ± 0.33 & 45.60 ± 20.98 & 2.89 ± 0.3 \\
Urdu & \textbf{19.21 ± 11.97} & 2.77 ± 0.42 & 19.93 ± 11.25 & \textbf{3.55 ± 0.27} & 19.38 ± 11.29 & 2.52 ± 0.28 & 19.85 ± 11.68 & 2.57 ± 0.30 & 31.34 ± 12.88 & 2.22 ± 0.3 & 47.09 ± 23.63 & 2.81 ± 0.29 \\
Vietnamese & 5.36 ± 8.31 & 3.20 ± 0.25 & 7.37 ± 10.92 & \textbf{3.21 ± 0.33} & \textbf{5.00 ± 8.73} & 2.76 ± 0.29 & 5.14 ± 7.28 & 2.89 ± 0.23 & 10.74 ± 9.99 & 2.5 ± 0.27 & 20.49 ± 22.68 & 2.72 ± 0.3 \\
Yoruba & 6.61 ± 7.48 & 3.17 ± 0.33 & 31.93 ± 16.40 & \textbf{3.36 ± 0.35} & \textbf{5.93 ± 7.26} & 2.96 ± 0.33 & 10.33 ± 10.00 & 2.95 ± 0.28 & 13.70 ± 10.40 & 2.77 ± 0.35 & 28.75 ± 21.50 & 3.02 ± 0.31 \\
\midrule
\textbf{Mean} & 18.74 ± 11.70 & 3.08 ± 0.28 & 26.86 ± 20.08 & \textbf{3.41 ± 0.09} & \textbf{16.95 ± 11.49} & 2.89 ± 0.26 & 21.50 ± 15.57 & 3.00 ± 0.20 & 31.13 ± 15.36 & 2.57 ± 0.26 & 44.51 ± 14.42 & 2.89 ± 0.21 \\
\bottomrule
\end{tabular}%
}
\caption{Per-language word error rate (WER, lower is better) and UTMOSv2 naturalness (higher is better) for each TTS system, computed on the first 500 utterances of the Open Bible test split. Values are reported as mean $\pm$ standard deviation across utterances. For each language and each metric, the best-performing system is in \textbf{bold}; ties are bolded jointly. Ndebele lacks an \texttt{omniASR\_LLM\_1B\_v2} language code and is reported with UTMOS only.}
\label{tab:wer-utmos-by-language}
\end{table*}

\section{Model Details and Training Configurations}
\label{sec:appendix_training_configuration}
EveryVoice, VITS, and F5-TTS were trained monolingually from scratch for each of the 37 \corpus{} languages under a matched update budget of 500,000 optimizer steps. Training was distributed across GH200, A100, H100, and L40 GPUs according to cluster availability, with the recipe and hyperparameters held fixed across devices; the number of GPUs per job was adjusted so that the total amount of data processed matched the 500,000-update budget. Table \ref{tab:training-config} summarizes model size and per-language wall-clock time; the paragraphs below describe the training recipe for each system. Reported durations are mean average times measured on a reference configuration of 2$\times$ NVIDIA L40S GPUs.

\begin{table}[h]
\centering
\small
\setlength{\tabcolsep}{4pt}
\renewcommand{\arraystretch}{1.05}
\begin{tabular}{@{}lcc@{}}
\toprule
\textbf{System} & \textbf{Parameters} & \textbf{Mean time (h)} \\
\midrule
EveryVoice & 18.2\,M & 25 \\
VITS       & 36.3\,M & 30 \\
F5-TTS     & 335.8\,M & 68 \\
\bottomrule
\end{tabular}
\caption{Trainable parameter counts and mean per-language wall-clock training time for the three from-scratch systems. Parameter counts follow the upstream implementations. Wall-clock times are reference measurements on 2$\times$ NVIDIA L40S GPUs; actual jobs used varying GPU types and counts as noted above.}
\label{tab:training-config}
\end{table}

\paragraph{EveryVoice.}
We use the EveryVoice toolkit with a non-autoregressive FastSpeech~2 acoustic model (18.2\,M parameters) and an iSTFTNet vocoder \citep{kaneko2022istftnetfastlightweightmelspectrogram}, a lightweight HiFi-GAN \citep{Kong2020HiFiGANGA} variant that replaces the final upsampling stages with an iSTFT. Training is multispeaker and monolingual: all utterances in a language share one acoustic model with learned speaker embeddings. Audio is resampled to 22,050 Hz. The full training pipeline comprises three stages: (1) feature extraction and filelist preparation over the full train split; (2) acoustic-model training from scratch for 500,000 optimizer updates (250,000 steps on 2 GPUs with DDP, batch size 32) (3) vocoder matching, in which mel spectrograms predicted by the trained acoustic model are used to fine-tune the pretrained universal iSTFTNet checkpoint for 100,000 steps (learning rate $10^{-5}$, batch size 32, DDP). Mean wall-clock time per language, including preprocessing, mel synthesis, and vocoder fine-tuning, was approximately 25 hours on the 2$\times$ L40S reference configuration.

\paragraph{VITS.} We train VITS end-to-end ($\sim$36.3\,M parameters) using the Coqui TTS implementation \footnote{\url{https://github.com/coqui-ai/tts}}. Open Bible waveforms are resampled to 22,050 Hz mono, each language is trained with learned multispeaker embeddings using multi-GPU distributed training (global batch size 64, 80 mel bins, hop length 256, window length 1024); on 2 GPUs this corresponds to 250,000 steps, equivalent to 500,000 single-GPU updates. Checkpoints are saved every 5,000 steps. Mean wall-clock time per language was approximately 30 hours on the 2$\times$ L40S reference configuration.

\paragraph{F5-TTS.}
We use the base F5-TTS configuration: a Diffusion Transformer (DiT) backbone with $\sim$335.8M parameters that maps text to a 100-band mel spectrogram at 24,000~Hz (hop length~256). A language-specific character vocabulary is built automatically from the training transcripts. The aligned \corpus{} split is preprocessed into a training configuration, and the number of training epochs is calibrated for the target GPU count so that optimization reaches 500,000 total updates. Training uses mixed \texttt{bf16} precision, a frame budget of 28{,}000 frames per GPU per step (at most 32 sequences), learning rate $7.5\times10^{-5}$, and 20{,}000 warmup updates. At inference, mel spectrograms are converted to waveform by the pretrained Vocos vocoder without per-language fine-tuning. Mean wall-clock time per language was approximately 68~hours on the 2$\times$ L40S reference configuration.

\section{Human Evaluation}
\label{sec:appendix_human_evaluation}

\subsection{Recruitment and compensation}
We recruited three self-identified native speakers for each of the ten languages. Hausa and Yoruba annotators were hired through direct community outreach; for the remaining eight languages (Haitian Creole, Hindi, Oromo, Shona, Swahili, Telugu, Turkish, and Vietnamese), annotators were hired through Upwork \footnote{\url{https://www.upwork.com/}}. Each annotator was paid 80 USD to complete the full task for their language. Annotators were instructed to use headphones in a quiet environment and were told that clips came from multiple TTS systems without revealing which model produced which sample.

\subsection{Annotation platform and task design}
\begin{figure}[h!]
    \centering
    \includegraphics[width=1.0\linewidth]{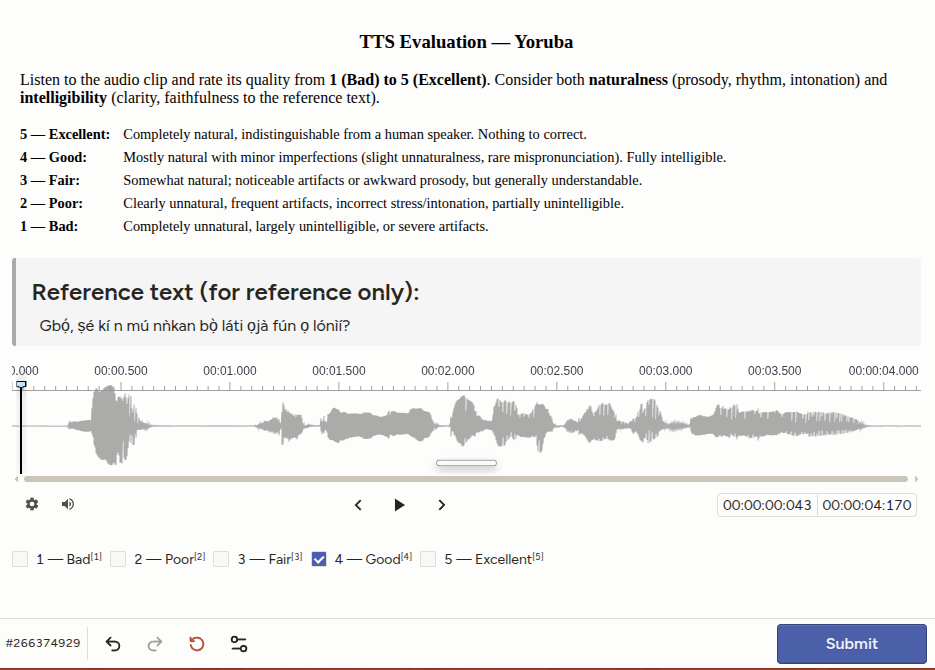}
    \caption{HumanSignal annotation interface for the listening study. Annotators listened to one clip at a time, read the reference transcript, and rated overall quality on a 1-5 MOS scale covering both naturalness and intelligibility.}
    \label{fig:HumanSignal-MOS-questions}
\end{figure}

Ratings were collected on HumanSignal\footnote{\url{https://humansignal.com/}}, a web-based annotation platform. We created one project per language; each project contained one task per audio clip, with model identity withheld from annotators. Each task presented one audio file together with the reference transcript; annotators assigned one overall score on a 1--5 scale from \emph{Bad} to \emph{Excellent}, judging both naturalness (prosody, rhythm, intonation) and intelligibility (clarity and faithfulness to the text). Figure~\ref{fig:HumanSignal-MOS-questions} shows the rating instructions displayed to annotators.

\begin{table}[h!]
\centering
\resizebox{\columnwidth}{!}{%
\begin{tabular}{@{}ccc@{}}
\toprule
\textbf{Dataset} & \textbf{Systems rated} & \textbf{Clips per language}           \\ \midrule
Open Bible       & Ground truth + 5 TTS   & 120                                   \\
FLEURS           & Ground truth + 5 TTS   & 120 (100 for Haitian Creole)          \\
BOUQuET          & 5 TTS only             & 100                                   \\ \midrule
\textbf{Total}   &                        & \textbf{340 (320 for Haitian Creole)} \\ \bottomrule
\end{tabular}%
}
\caption{Human-evaluation stimulus counts. Each dataset contributes 20 utterances; clip totals multiply utterances by the number of rated systems available for that dataset.}
\label{tab:human-eval-stimuli}
\end{table}

Every annotator rated all clips in their language panel. Table~\ref{tab:human-eval-stimuli} summarizes the stimulus counts. For each dataset we fixed 20 utterance indices and generated all systems on those texts. Open~Bible and FLEURS clips include ground-truth recordings alongside the five TTS systems (six outputs per utterance). BOUQuET provides no reference speech, so only the five synthetic systems were rated (five outputs per utterance). Haitian Creole has no FLEURS ground-truth audio, reducing its FLEURS block from 120 to 100 clips.


\end{document}